\documentclass[lettersize,journal]{IEEEtran}
\usepackage{amsmath,amsfonts}
\usepackage{algorithmic}
\usepackage{algorithm}
\usepackage{array}
\usepackage[caption=false,font=normalsize,labelfont=sf,textfont=sf]{subfig}
\usepackage{textcomp}
\usepackage{stfloats}
\usepackage{url}
\usepackage{verbatim}
\usepackage{graphicx}
\usepackage{cite}
\usepackage{enumitem}
\usepackage{multirow}
\usepackage{booktabs}
\usepackage{makecell}
\usepackage{xcolor}

\begin{document}

\title{A Multi-Level Similarity Approach for Single-View Object Grasping: Matching, Planning, and Fine-Tuning}

\author{Hao Chen, Takuya Kiyokawa, Zhengtao Hu, Weiwei Wan, and Kensuke Harada
\thanks{All authors are with the Department of Systems Innovation, Graduate School of Engineering Science, Osaka University, Toyonaka, Osaka 560-8531, Japan \textit{(Corresponding author: Hao Chen, e-mail: hchenrsjp@gmail.com)}.}
\thanks{Zhengtao Hu is also with the School of Mechatronic Engineering and Automation, Shanghai University, Shanghai 200444, China.}
\thanks{Kensuke Harada is also with the National Institute of Advanced Industrial Science and Technology (AIST), Tokyo 100-8921, Japan.}}

\markboth{}%
\IEEEpubid{}

\maketitle

\begin{abstract}
Grasping unknown objects from a single view has remained a challenging topic in robotics due to the uncertainty of partial observation. Recent advances in large-scale models have led to benchmark solutions such as GraspNet-1Billion. However, such learning-based approaches still face a critical limitation in performance robustness for their sensitivity to sensing noise and environmental changes. To address this bottleneck in achieving highly generalized grasping, we abandon the traditional learning framework and introduce a new perspective: similarity matching, where similar known objects are utilized to guide the grasping of unknown target objects. We newly propose a method that robustly achieves unknown-object grasping from a single viewpoint through three key steps: 1) Leverage the visual features of the observed object to perform similarity matching with an existing database containing various object models, identifying potential candidates with high similarity; 2) Use the candidate models with pre-existing grasping knowledge to plan imitative grasps for the unknown target object; 3) Optimize the grasp quality through a local fine-tuning process. To address the uncertainty caused by partial and noisy observation, we propose a multi-level similarity matching framework that integrates semantic, geometric, and dimensional features for comprehensive evaluation. Especially, we introduce a novel point cloud geometric descriptor, the C-FPFH descriptor, which facilitates accurate similarity assessment between partial point clouds of observed objects and complete point clouds of database models. In addition, we incorporate the use of large language models, introduce the semi-oriented bounding box, and develop a novel point cloud registration approach based on plane detection to enhance matching accuracy under single-view conditions. Real-world experiments demonstrate that our proposed method significantly outperforms existing benchmarks in grasping a wide variety of unknown objects in both isolated and cluttered scenarios, showcasing exceptional robustness across varying object types and operating environments.
\end{abstract}


\section{Introduction}
\IEEEPARstart{B}{oth} industrial and service robots are required to deal with a wide range of objects with diverse categories, shapes and arrangements. In traditional robotic systems, detailed properties of each object have to be known and then specific actions can be designed for each manipulation task. However, this process becomes increasingly labor-intensive as the number of object types grows, while remaining ineffective for previously unseen objects. Consequently, there is a pressing need for highly dexterous robotic systems capable of handling a wide variety of novel objects without prior information.

In the last decade, the booming development of vision~technology and deep learning has led to many outstanding works in the field of novel object grasping \cite{Newbury}. Representative studies among them use depth images \cite{Mahler,Morrison}, RGBD images \cite{Yan,Zeng} or point clouds \cite{Pas,Mousavian} as input representations to train neural networks for grasp detection and evaluation on unseen objects. While achieving notable performance, they have limitations in the grasping direction (e.g. only top-down grasping) or rely on high-precision visual features for accurate grasp inference. For improvement, more recent studies employ scene representations \cite{Breyer,Jiang} and large-scale datasets \cite{Fang,Vuong} to provide efficient and adequate training for more generalized grasping systems. However, they still suffer from the disadvantages of high training costs and high sensitivity to sensing noise and environmental changes. Therefore, we recognize that a new perspective beyond traditional learning frameworks is needed to achieve a higher standard of general object grasping, leading us to explore the idea of similarity matching \cite{Goldfeder,Mitrevski}.

\begin{figure}[t]
    \centering
    \includegraphics[width=\linewidth]{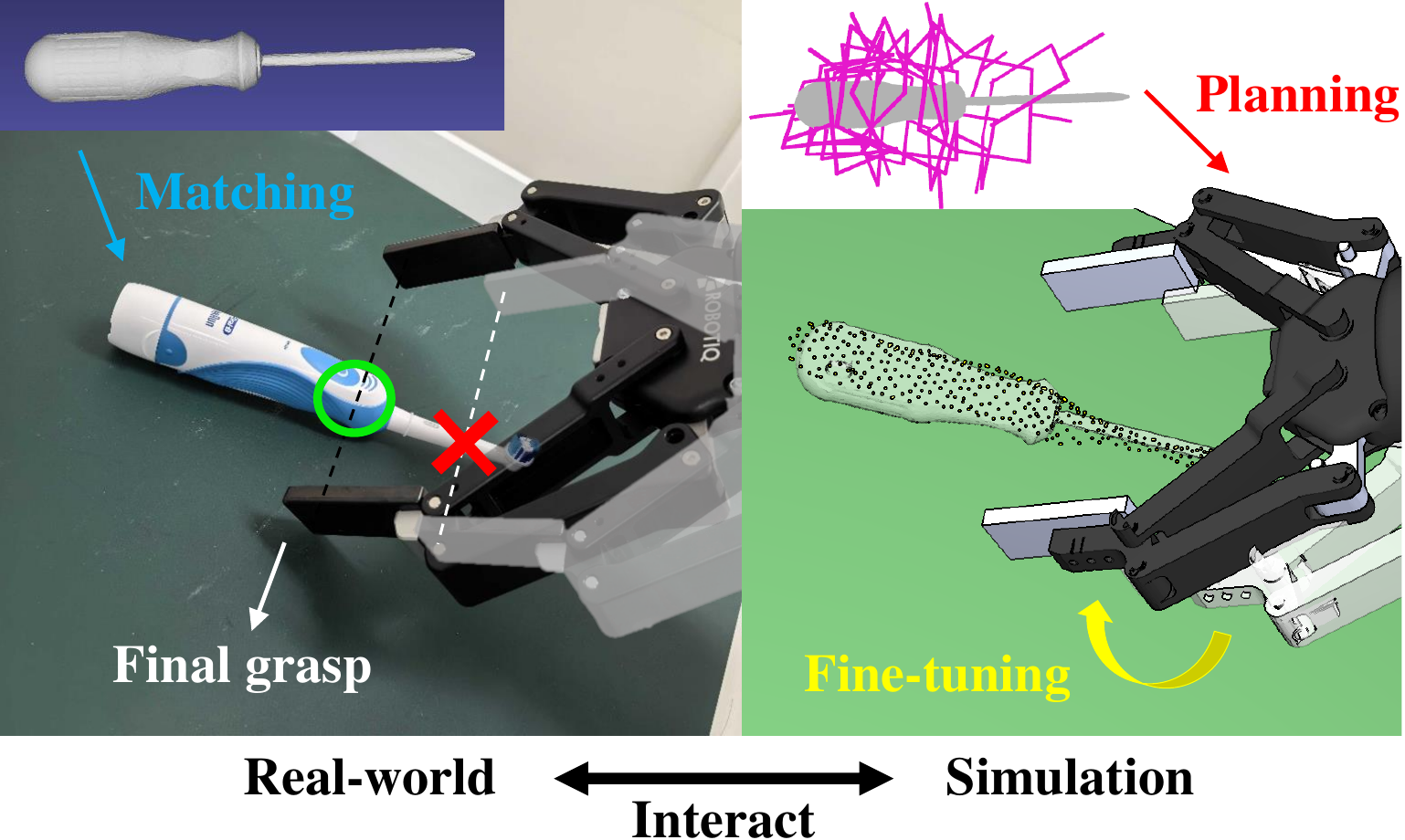}
    \caption{An example of grasping a novel object using our three-step method. First, visual features of the target object obtained in the real world are used for similarity matching with existing database models. Then, grasp planning and fine-tuning are performed in simulation based on the matching results. Finally, the optimized grasp is executed in the real world to achieve the task.}
    \label{img:1}
\end{figure}

A recent enlightening work \cite{Chen} introduces a score-based similarity evaluation framework that assesses object similarity from both semantic and geometric aspects. This method effectively transfers grasping knowledge from similar known objects to enable efficient grasping of unknown target objects. However, it requires multi-view observations and exhibits instability as the variety of target objects increases, primarily due to the difficulty in balancing the quantification of semantic and geometric similarity. Our study significantly advances this work by tackling the challenge of single-view object grasping and developing a more efficient similarity matching approach that fully leverages available visual features. We introduce a novel multi-level matching framework that independently evaluates object similarity across semantics, geometry, and dimensions, avoiding the balancing issues of using a composite scoring function. Notably, we propose the C-FPFH (Clustered Fast Point Feature Histogram) descriptor for geometric matching, which, to the best of our knowledge, is the first geometric descriptor capable of accurately evaluating the similarity between partial and complete point clouds of non-identical objects, demonstrating exceptional effectiveness in handling occlusions. In addition to accurate matching, we incorporate a stability-aware fine-tuning process to optimize the quality of imitative grasps generated from similar references, providing an auxiliary guarantee for achieving robust grasping.

Through extensive real-world experiments with a diverse range of novel objects in both isolated and cluttered scenes, we show that our method, using a small database of fewer than 100 object models, significantly outperforms state-of-the-art (SOTA) benchmarks across key metrics including accuracy, efficiency, and generalization. Fig. \ref{img:1} illustrates an example of grasping a novel object (a toothbrush) using our proposed method. Through visual detection and similarity matching, a screwdriver model with grasping knowledge is identified from an existing database to plan imitative grasps for the toothbrush. A subsequent fine-tuning process is then applied to optimize the grasp quality by positional adjustment. The optimized grasp is finally executed to complete the task. The entire system operates through a seamless interaction between the real world and the simulation.

Our main contributions can be summarized as:

\begin{itemize}

\item We propose a multi-level similarity matching approach that integrates semantic, geometric, and dimensional features to efficiently identify potential similar candidates from an existing database for the unknown target object. 

\item In geometric matching, we introduce the C-FPFH descriptor, a novel feature descriptor designed to accurately assess similarity between partial and complete point clouds.

\item We develop several new methods to enhance the accuracy of similarity matching and grasp planning by exploiting and improving existing techniques such as Large Lanuage Models (LLMs), Oriented Bounding Boxes (OBBs), and plane detection in point clouds. 

\item We implement a two-stage fine-tuning process after generating imitative grasps to optimize grasp stability based on the local features of observable contact points.

\end{itemize}

\section{Related Work}
Most of the existing research focuses on using learning-based approaches to address the challenge of novel object grasping. Some recent works also achieve notable results using purely geometric analysis. Additionally, a few studies explore the use of similarity matching; however, its potential remains largely underutilized and deserves further exploration.

\textbf{Learning-based approaches.} In the 2010s, major advances in Convolutional Neural Networks (CNNs) enabled robotic vision to achieve unprecedented performance in novel object grasping, spawning outstanding works such as Dex-Net \cite{Mahler}, GG-CNN \cite{Morrison}, and GPD \cite{Pas}. They leverage a large number of depth images or point clouds with labeled grasp poses to train CNN models capable of predicting high-quality grasp poses for unseen objects. DGGN \cite{Yan} first uses RGBD inputs to train a two-stage network regressing grasps from reconstructed 3D scenes. PointNetGPD \cite{Liang} enhances the performance of GPD by integrating the architecture of PointNet \cite{Charles} into an end-to-end grasp evaluation network. QT-Opt \cite{Kalashnikov} represents an attempt at leveraging reinforcement learning for grasp generalization, achieving high grasp success rates on unseen objects through large-scale self-supervised training. In the 2020s, a broader range of methods beyond traditional CNNs has been explored. VGN \cite{Breyer} and GIGA \cite{Jiang} employ Truncated Signed Distance Field (TSDF) representations to efficiently learn grasp detection in cluttered scenes. 3DSGrasp \cite{Mohammadi} and SCARP \cite{Sen} perform shape completion on partial point clouds to enhance the performance of single-view grasping. GraspNet \cite{Fang} and Grasp-Anything~\cite{Vuong} construct large-scale grasp datasets and use them to train high-performance models for general object grasping. HGGD \cite{Siang} provides new insights into generating dense grasps in real-time by utilizing both global and local features of objects in clutter. While these methods perform well in their specific tasks, they share several common limitations: 1) high training cost; 2) high sensitivity to sensing noise and environmental changes; and 3) low error traceability due to complex learning architectures. In contrast, our proposed method is training-free, robust to varying conditions, and fully error-traceable through a simple framework.

\textbf{Analysis-based approaches.} A few recent studies try to generate high-quality grasps directly from object point clouds by geometric analysis. Adjigble et al. \cite{Adjigble} leverage \textit{zero-moment shift features} \cite{Clarenz} to evaluate the local geometric similarity between object surfaces and gripper surfaces, enabling the selection of grasp positions with the highest probability of success. Wu et al. \cite{Wu} detect \textit{hidden superquadrics} \cite{Liu} from partial object point clouds to generate and filter reliable grasp candidates through a multi-metric evaluation. Wang et al. \cite{Wang} propose \textit{visible point-cloud} to efficiently exclude unsafe grasps and determine the optimal grasp pose from a partial view. These methods show the possibility of generalized grasping without model training. However, they require high-precision visual features to achieve good results, whereas our method remains effective even with sparse and noisy visual inputs.

\textbf{Similarity-based approaches.}
The Columbia Grasp Data-base \cite{Goldfeder} first introduced similarity matching in novel object grasping. They leverage similar database models with precomputed grasps to achieve imitative grasping of previously unseen objects. However, their similarity computation requires prior 3D scanning of the target object, leading to inefficient task completion. Herzog et al. \cite{Herzog} develop a template-based grasp planning algorithm that generalizes demonstrated grasps to novel objects with similar local geometry. However, their performance depends heavily on the number and type of grasp templates. Two recent studies \cite{Mitrevski, Chen} utilize ontological classification and scoring functions to identify similar objects with grasping knowledge for guiding the grasping of novel objects. Although these methods perform inconsistently due to their reliance on existing knowledge, their efforts highlight that the potential of using similarity matching has been underestimated, inspiring us to further explore this direction.

\begin{figure*}[t]
    \centering
    \includegraphics[width=\linewidth]{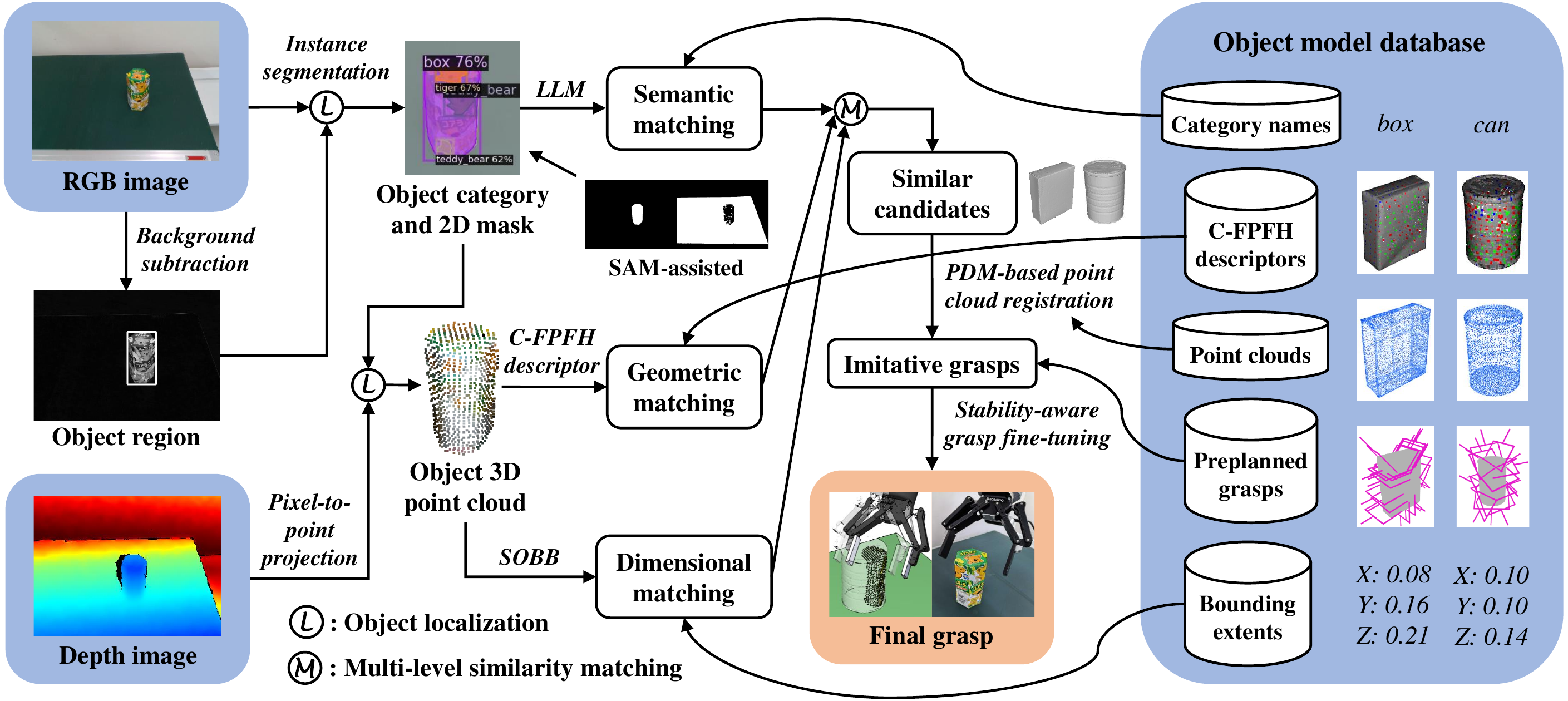}
    \caption{Overview of our proposed approach. The system inputs (shown in blue boxes) include a single-view RGBD image and an existing object model database. A background image without target objects is taken beforehand. The output (shown in the orange box) is an optimized grasping action for execution.}
    \label{img:2}
\end{figure*}

\section{Methods}
\subsection{Overview}
The core idea of our approach is to leverage visual features from single-view observations to identify similar references from an existing database of known object models to guide the grasping of unknown target objects. The main difficulty lies in achieving accurate similarity matching and robust grasp planning in the presence of large visual uncertainty.

Fig. \ref{img:2} illustrates an overview of our methodology. The first step involves extracting object features from single-view RGBD inputs using segmentation models, including category information and 3D point clouds. The second step applies a multi-level similarity matching approach with an object model database consisting of the following components: 

1) \textit{Category names}, used for semantic matching with the detected object category through LLM assistance; 

2) \textit{C-FPFH descriptors}, used for geometric matching with the object 3D point cloud via feature comparison; 

3) \textit{Bounding extents}, used for dimensional matching with the target object by utilizing a novel type of OBB.

Based on the three levels of matching, we filter a list of similar candidates from the database that are most likely to exhibit high similarity to the target object. For these candidates, we further rank their similarity by leveraging another database component: 4) \textit{Point clouds}, to perform point cloud registration with the object point cloud using a plane-detection-based approach. Based on the registration results, we begin with the most similar candidate and transfer its existing grasping knowledge from the database: 5) \textit{Preplanned grasps}, to generate imitative grasps for the target object. 

In the final step, all generated grasps undergo a two-stage fine-tuning process to optimize stability, considering the local features of observable contact points.

\subsection{Single-View Object Recognition}
Our goal is to achieve single-view object grasping under the uncertainty of sparse and noisy visual inputs. To achieve this, we use a consumer-grade 3D camera mounted on the robot end effector for object detection. We first capture a background image without the target object, and then use it as a reference for background subtraction by performing gray-scale differencing with the image containing the object to identify the region where the object is located. Based on this approximate localization, we apply instance segmentation to the RGB input using a SOTA pre-trained model \cite{Zhou} and extract only the results related to the target object, including the object category and its 2D mask. 

To remove redundant detection results such as the patterns on the object surface, we also examine the inclusion relationships among different results and retain only the one with the largest coverage within the object region. In cases where the target object fails to be detected (confirmed by comparing the object region with all detection results) due to irregular observation angles or occlusions, we integrate a more powerful segmentation model, SAM \cite{Kirillov}, to acquire the object mask without category information. Based on this mask, we extract the pixels containing the target object from the depth input and project them into 3D space to obtain the object point cloud. 

It should be noted that the object region identified by background subtraction cannot replace the 2D mask acquired by instance segmentation since it lacks pixel-level accuracy and is highly susceptible to lighting variations. The object category and 3D point cloud obtained during the visual detection process are key features for achieving similarity matching.

\subsection{Multi-Level Similarity Matching}
Existing learning-based grasp planning approaches struggle to maintain their optimal performance under varying sensing conditions. To address this issue, we avoid directly regressing grasps from visual inputs. Instead, we utilize existing similar objects with reliable grasping knowledge to guide the grasping of novel objects, which is robust to noise and environmental changes through an appropriate matching framework. 

Based on this consideration, we propose a multi-level similarity matching approach that leverages the visual features of the target object to identify reference object models from an existing database, selects candidates from semantic, geometric, and dimensional perspectives, respectively, and finally synthesizes the results to determine an optimal selection of similar candidates. Compared with score-based methods \cite{Chen}, the multi-level matching framework exhibits greater robustness in similarity evaluation by separately assessing potential candidates from each perspective, thus avoiding influences across different perspectives. The implementation details of each matching are illustrated in the following sections.

\subsection{LLM-Assisted Semantic Matching}
The object category obtained in visual detection can be used to identify similar candidates from the database at the semantic level. A common approach for evaluating semantic similarity between two categories is to use word embedding models such as Word2Vec \cite{Mikolov} and GloVe \cite{Pennington} to obtain scores representing their cosine similarity (e.g., \textit{bottle} has a high score with \textit{box}, but a low score with \textit{car}). However, this method has several significant drawbacks that can mislead the results of similarity matching: 1) The scores only reflect word similarity in linguistics, not relevance in robotic grasping; 2) The scores vary significantly across different pairs of similar categories; 3) Different meanings of a polysemous word cannot be distinguished. Fortunately, with advancements in LLMs, we address these issues by integrating the GPT-4o model~\cite{OpenAI} for similarity evaluation, which provides comprehensive knowledge of object similarity specific to robotic grasping and a better understanding of user intent, as implemented below:

\textbf{Prompt}: Which objects in [\textit{bottle, box, cup, mug, apple, hammer}] are likely to be similar to a \{\textit{soda\_can}\} in terms of robotic grasping? Please only answer the category names.

\textbf{Answer}: Bottle, cup, mug.

In the first square bracket, we input all category names contained in the database, simplified to single or compound words that indicate object identities without descriptive terms (e.g., for objects in the YCB dataset \cite{Calli}, 051\_large\_clamp $\to$ clamp, 053\_mini\_soccer\_ball $\to$ soccer\_ball). In the second curly bracket, we input the detected object category without any simplification, as the descriptive terms in the detection results, e.g., \textit{mouse\_(computer\_equipment)}, help to better clarify the target object. From the GPT answer, we can easily select candidate models with corresponding category names that are potentially similar to the target object from a semantic perspective. In case that the object fails to be detected and no category information is available, we skip semantic matching and only consider similarity from other perspectives.

\subsection{C-FPFH-Based Geometric Matching}\label{FPFH}
The object point cloud obtained in visual detection can be used to represent the geometric properties of the target object, enabling the identification of similar candidate models at the geometric level. However, incomplete point clouds from single-view observations introduce significant uncertainty due to large unseen regions and sensing noise, making it extremely difficult to accurately assess the similarity between an observed partial point cloud of an unknown object and a complete point cloud of a database model. To address this issue, rather than exploring the global geometric similarity between partial and complete point clouds, we find it more effective to extract their local geometric features and leverage feature correspondences to represent their similarity. To achieve this, we adopt a point cloud feature extractor, the FPFH descriptor \cite{Rusu}, which describes the local geometry of each point in a point cloud using a 33-dimensional vector showcasing the distribution of neighboring normal orientations. It has the advantage of rotational invariance and is suitable for our task where the camera observation pose is uncertain. Utilizing its principle, we propose a novel point cloud geometric descriptor, the C-FPFH descriptor, which aggregates and classifies local geometric features to distill the essential information in a point cloud, enabling accurate similarity evaluation regardless of point cloud completeness.

\begin{figure}[t]
    \centering
    \includegraphics[width=\linewidth]{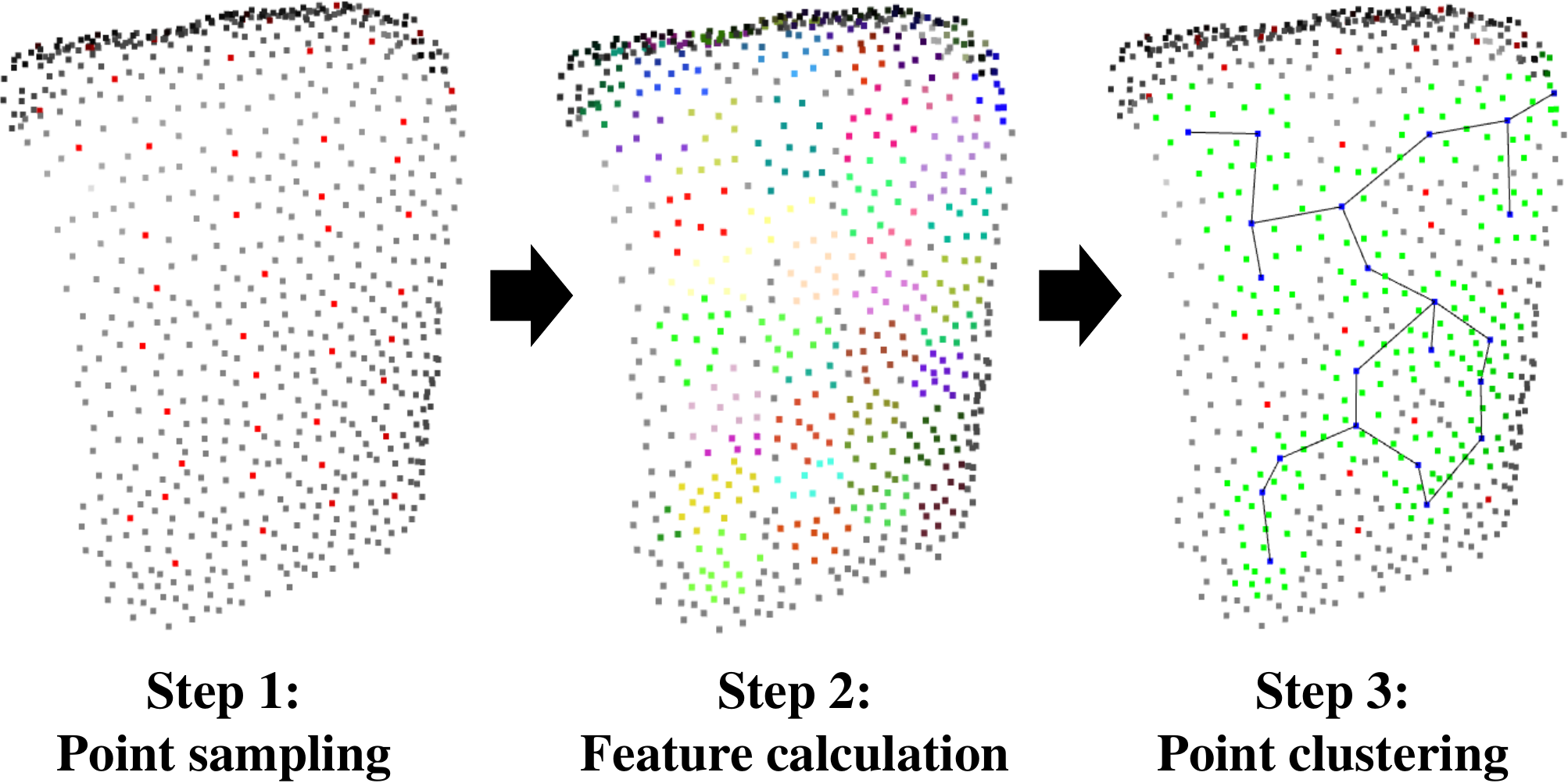}
    \caption{Three-step generation of C-FPFH descriptors for object point clouds. Due to sensing noise, point features may occasionally vary across surfaces with similar geometry; however, such variations are acceptable during geometric matching, facilitated by our discrete sampling and consecutive clustering methods.}
    \label{img:3}
\end{figure}

As shown in Fig. \ref{img:3}, generating a C-FPFH descriptor for a point cloud involves three steps: 1) Uniformly sample points using a voxel grid filter of appropriate size (1.5 cm in our task) and exclude edge points where the estimated normals probably deviate from the truth due to the uncertainty of unseen regions. 2) For each sampled point, compute its FPFH descriptor with neighboring points within an appropriate local area (a sphere with a radius of 1 cm in our task) to obtain a 33-dimensional vector describing its local geometry. To minimize the effect of sensing noise, instead of using all the values in the vector, we extract only the index numbers of the two most dominant of the 33 vector components as a feature pair to represent the main characteristics of a sampled point. The rationale for employing two principal components rather than any other number is detailed in Appendix \ref{details}. We aggregate the feature calculation results of all sampled points as: $\{(f_{11},f_{12}):n_1, (f_{21},f_{22}):n_2, ...\}$, where $(f_{k1},f_{k2}):n_k$ is an unordered feature pair with its number of occurrences in the point cloud. Based on this aggregation result, we develop the first metric for similarity evaluation between a partial point cloud $p$ and a complete point cloud $c$, called QS (quantitative similarity), which is calculated as:
\begin{equation}
    \label{eqn:1}
    {\rm QS}=\frac{{\textstyle \sum_{i=1}^{m}}\min(^{p}n_{i},\\^{c}n_{i})}{{\textstyle \sum_{i=1}^{m}}^{p}n_{i}} 
\end{equation}
where $m$ is the number of types of feature pairs in $p$, $^{p}n_{i}$ and $^{c}n_{i}$ are the number of times the same feature pair $(f_{i1},f_{i2})$ occurs in $p$ and $c$, respectively ($^{p}n_{i}$ is always greater than 0, while $^{c}n_{i}$ can be 0). A large QS indicates that $c$ contains most of the features in $p$, in which case it is highly probable that they are similarly shaped objects. However, this metric has limitations as it only evaluates the overlap rate of identical features without considering their spatial distribution. As a result, two dissimilar point clouds containing a large number of discrete identical features may be evaluated as highly similar. Therefore, an additional metric that captures the similarity of feature distribution is essential to complement the evaluation.

\begin{figure*}[t]
    \centering
    \includegraphics[width=0.9\linewidth]{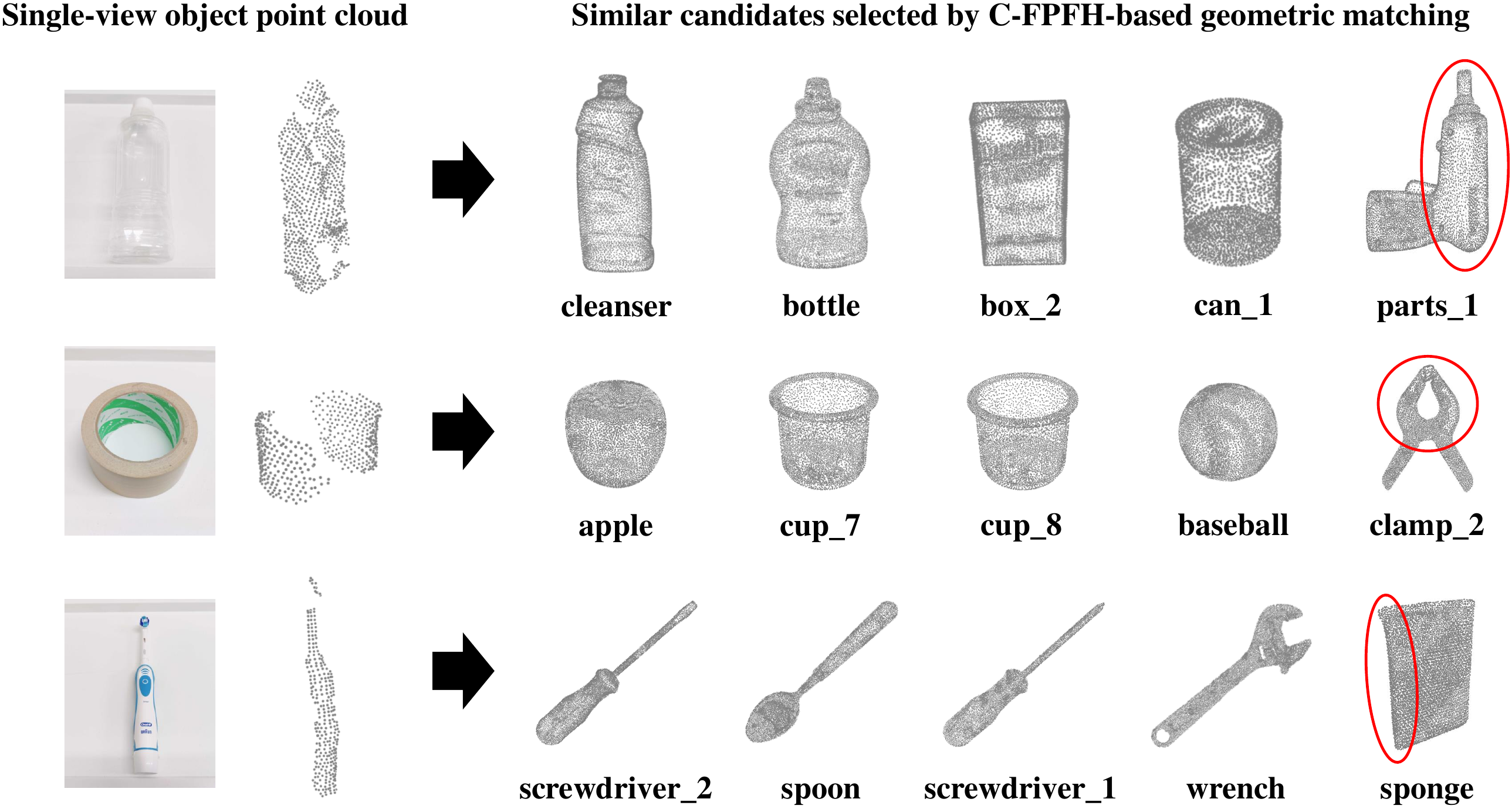}
    \caption{Test results of using C-FPFH-based geometric matching to identify similar models from single-view object point clouds. All candidate models are originally from the YCB dataset, renamed with simplified category names and index numbers for cases where multiple models exist within a single category.}
    \label{img:4} 
\end{figure*}

3) Considering that the surface features in $p$ are not as complete as those in $c$, the comparison of feature distribution needs to be performed locally rather than globally. To achieve this, we incorporate the idea of the CVFH descriptor \cite{Aldoma} to cluster all sampled points as follows: I) Take an unclustered sampled point as a seed, search its surrounding area (a sphere with a radius slightly larger than the sampling voxel size, 2 cm in our task) to identify other sampled points with the same feature pair and similar normal orientations (differing by no more than 20 degrees in our task); II) If such points are found, connect them as consecutive points and take them as new seeds, continuing the search until no more consecutive points are found; III) Repeat I and II. In this way, each sampled point can be assigned to a unique cluster. All clusters are confined to a single surface without crossing the edges, ensuring that the feature distribution is evaluated in a localized manner. For each cluster with more than two points, we apply principal component analysis (PCA) to obtain three normalized singular values as a representation of its spatial distribution. We aggregate the distribution analysis results for clusters with the same feature pair as: $\{(f_{k1},f_{k2}):d_{k1}=[\sigma_{11},\sigma_{12},\sigma_{13}]_k, d_{k2}=[\sigma_{21},\sigma_{22},\sigma_{23}]_k, ...\}$. For $p$, we only focus on the distribution of the main features (the cluster containing the highest number of consecutive points with the most frequently occurring feature pair) and obtain its PCA result as $^pd_s=[\sigma_{1},\sigma_{2},\sigma_{3}]_s$, where $s$ denotes that $(f_{s1},f_{s2})$ is the main feature pair of $p$. We ignore unimportant clusters and feature pairs for the presence of sensing noise. For $c$, we search among the clusters of $(f_{s1},f_{s2})$ to identify any PCA result that is close to $^pd_s$. When such a result exists, we assume that $c$ contains the main point cluster of $p$. Based on this principle, we develop the second metric for similarity evaluation, called DS (distributional similarity), which is calculated as:
\begin{equation}
    \label{eqn:2}
    {\rm DS}=\min_{1\le j\le l}\left \|^pd_s-\\^cd_{sj}\right \| 
\end{equation}
where $l$ is the number of clusters with feature pair $(f_{s1},f_{s2})$ in $c$. A small DS indicates that $c$ has a region very similar to the main part of $p$.

The aggregation results of feature calculation and distribution analysis form the C-FPFH descriptor. To save computation time in similarity matching, we pre-generate C-FPFH descriptors for all complete point clouds in the database. This allows us to compute only the descriptor for the partial point cloud of the target object during real-time processing. By setting appropriate thresholds for QS and DS, we can determine the range of similar candidate models from a geometric perspective. Empirically, we assume that when ${\rm QS}>0.9$ and ${\rm DS}<0.1$, $p$ has high geometric similarity to $c$ and the model is selected as a candidate. These two thresholds can be adjusted to increase or decrease the number of candidates; however, setting them too high or too low can negatively impact the matching results.

Fig. \ref{img:4} showcases several test results of using C-FPFH-based geometric matching to identify similar candidate models from single-view object point clouds. The results indicate that most of the selected models demonstrate overall similarity to the target objects. However, some candidates display only partial similarity (marked with red circles in Fig. \ref{img:4}) due to the inclusion relationship between $p$ and $c$, revealing a potential limitation of using local features for similarity evaluation between partial and complete point clouds.

\subsection{SOBB-Guided Dimensional Matching}\label{SOBB}
\begin{figure}[t]
    \centering
    \includegraphics[width=\linewidth]{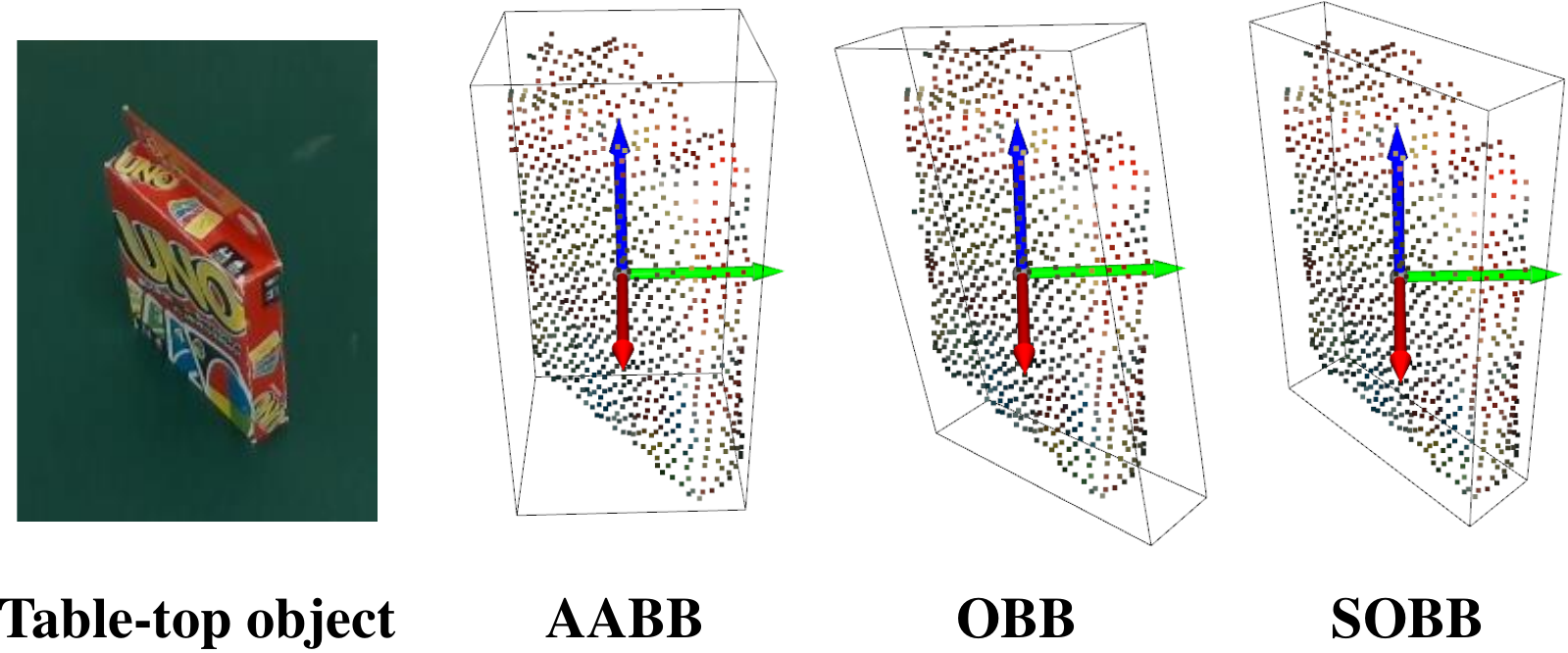}
    \caption{Comparison of different types of bounding boxes applied to a single-view partial point cloud obtained from a table-top object.}
    \label{img:5} 
\end{figure}

To address the limitation of partial similarity in geometric matching, we introduce a third perspective for similarity evaluation: dimensional matching, which assesses the size similarity between the target object and database models by leveraging their dimensional features. For size evaluation of object models or point clouds, 3D bounding boxes such as the axis-aligned bounding box (AABB) and the oriented bounding box (OBB) are commonly used to represent dimensions including length, width, and height through the three bounding extents. However, in our task, neither AABB nor OBB can accurately represent the dimensions of the target object from single-view partial point clouds due to the uncertainty of object poses and large unseen regions, as shown in Fig. \ref{img:5}. To address this issue, we develop a new type of bounding box, called semi-oriented bounding box (SOBB), which fixes one direction of the OBB by aligning it with the normal vector of the plane on which the target object is placed.

The generation of the SOBB for an object point cloud involves the following steps: 1) In the robot coordinate system, obtain the normal vector $\Vec{n}$ of the plane where the object is placed (for table-top objects in our task, $\Vec{n}=\left \langle 0,0,1 \right \rangle $); 2) Project all points onto a plane perpendicular to $\Vec{n}$ and passing through the origin; 3) Downsample the projected points using a grid filter to achieve a uniform density; 4) Apply PCA to the downsampled points to extract the two principal components, $\Vec{u}$ and $\Vec{v}$, which represent the 2D distribution of the points within the plane; 5) Generate a new coordinate system using normalized $\Vec{n}$, $\Vec{u}$, and $\Vec{v}$ as the orthogonal basis vectors, and compute the AABB of the point cloud in this new coordinate system. By transforming this AABB back to the original coordinate system, we obtain the target SOBB, which accurately represents the dimensions of the target object.

For evaluation of size similarity, we pre-compute the bounding extents $(x_i,y_i,z_i)$ for all database models by aligning their original poses with the axis orientations and obtaining their corresponding AABBs. During real-time processing, we only need to compute the SOBB extents $(x,y,z)$ of the object point cloud and use its dimensional differences with the database models to represent their size similarity (SS) as follows:
\begin{equation}
    \label{eqn:3}
    {\rm SS}=\left \| {\rm sort}(x,y,z)-{\rm sort}(x_i,y_i,z_i)_{1\le i\le h} \right \|
\end{equation}
where $h$ is the number of database models, ${\rm sort}()$ is a function that arranges values in descending order. All calculations are performed in meters. Empirically, we assume that when ${\rm SS}<0.1$, the model is similar in size to the target object and can be selected as a candidate at the dimensional level. This threshold is also adjustable but should remain within an appropriate range, similar to QS and DS.

\subsection{Candidate Model Selection}\label{selection}
The above implementations indicate that each individual level of similarity matching has its limitations. For example, semantic and dimensional matching only provide a coarse identification of potential similar candidates without investigating into detailed object features, whereas geometric matching leverages object features thoroughly but is prone to partial similarity results. Therefore, we propose that the most effective approach for candidate model selection is to use multi-level similarity matching, following the principle that a model identified as similar across more perspectives is more likely to resemble the target object. Based on this principle, the final candidate models are selected in the following order: 
\begin{enumerate}[label=(\roman*)]
\item{Models similar in all three perspectives; }
\item{Models similar in two perspectives (if no models meet the criteria of i); }
\item{Models similar in one perspective (if no models meet the criteria of i and ii); }
\item{Non-similar models (if no models meet the criteria of i, ii, and iii). }
\end{enumerate}

This approach ensures that all perspectives are considered equally, thereby avoiding bias or balancing issues.

\subsection{PDM-Based Point Cloud Registration}
After selecting candidate models through similarity matching, two issues remain to be solved: 1) Which candidate model should be prioritized as the reference for object grasping? 2) How to transfer the grasping knowledge from a similar database model to the unknown target object? To address them, we perform point cloud registration between the partial point cloud $p$ of the target object and the complete point clouds $c$ of all candidate models to obtain their fitness scores and transformation matrices, which can be used to determine the model priority and transfer grasping knowledge, respectively.

\begin{figure}[t]
    \centering
    \includegraphics[width=0.9\linewidth]{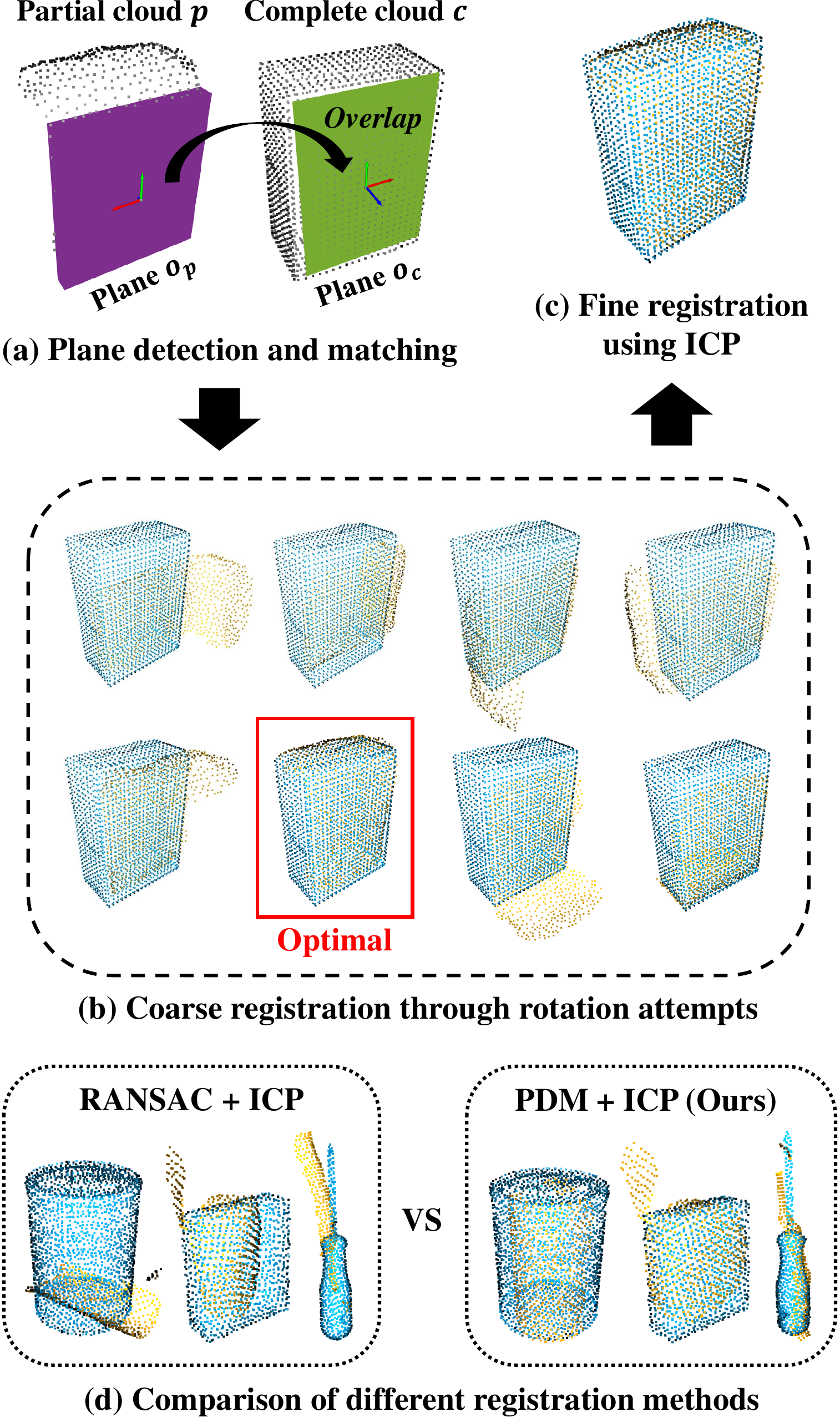}
    \caption{(a)-(c): Three steps to perform PDM-based point cloud registration between partial (yellow) and complete (blue) point clouds. (d): Comparison between a traditional registration method and our proposed method.}
    \label{img:6}
\end{figure}

Traditional point cloud registration methods employ algorithms such as RANSAC \cite{RANSAC} and ICP \cite{ICP}  to achieve coarse-to-fine registration between point clouds of \textbf{identical objects}. However, our task copes with point clouds of \textbf{similar objects}, where the inherent randomness of the RANSAC algorithm can lead to suboptimal initial alignments and unreliable registration results (see Fig. \ref{img:6}d). To solve this issue, we adopt a plane detection method \cite{Araújo} to improve the coarse registration process. As shown in Fig. \ref{img:6}a-c, we first detect the largest plane $o_p$ in $p$ and the plane $o_c$ in $c$ with the area closest to $o_p$. Then, we align $p$ with $c$ by overlapping the two coordinate systems located at the centers of $o_p$ and $o_c$. However, this overlapping does not necessarily result in an optimal alignment between $p$ and $c$ due to the uncertainty of the axis orientation. To address this, in the next step, we apply various rotations around the plane center to identify the optimal rotation that minimizes the distance between $p$ and $c$, and use it as the coarse registration result. Based on this result, we finally employ the ICP algorithm for fine registration. The advantage of using plane detection and matching (PDM) is that it ensures a large overlap area between the two registered point clouds, regardless of their similarity and completeness. Additionally, the results are consistent across different trials, eliminating randomness. A performance comparison of RANSAC + ICP versus PDM + ICP is shown in Fig. \ref{img:6}d. In most cases, PDM + ICP achieves better results. However, in scenarios where no plane is detected in $p$ due to complex object geometries or sensing noise, PDM becomes inapplicable and we switch the coarse registration method to RANSAC. In practice, such switches are rarely observed, occurring in fewer than 10\% of cases—typically involving transparent or small objects whose point clouds are more prone to errors, thereby not significantly impacting grasp planning performance.

\subsection{Imitative Grasp Planning}\label{ranking}
Each point cloud registration result includes a fitness score representing the overlap rate between the two registered point clouds and a transformation matrix defining their relative pose. Based on the fitness score, we rank the candidate models from highest to lowest priority and use them sequentially during grasp planning. Based on the transformation matrix, we transfer the grasping knowledge from candidate models to the target object through the following steps: 

1) We preplan hundreds of robust grasps for each database model using a mesh segmentation approach \cite{Wan}; 2) For the candidate model with the highest priority, we apply the transformation matrix to its preplanned grasps to generate imitative grasps on the target object; 3) These grasps are then evaluated in a simulation environment \cite{wrs} (also used for the subsequent fine-tuning process), where IK computations are achieved by IKFast \cite{Diankov} and collision detection are performed by reconstructing a model of the target object from its observed point cloud using the ball-pivoting algorithm \cite{Bernardini}, in order to exclude unreachable and collided grasps; 4) For the remaining grasps, we let the robot reach each grasp pose in the simulation and perform further filtering as follows: Assuming that a cube model filling the gripper closure region is $\mathbb{C}$, the gripper model is $\mathbb{G}$, and the object point cloud is $\mathbb{P}$, a grasp is not on the object when $\mathbb{C}\cap \mathbb{P} = \emptyset$, and a grasp collides with the object when $\mathbb{G}\cap \mathbb{P} \ne \emptyset$. Such infeasible grasps occur due to the inherent differences between the similar model and the real object, and are excluded during planning; 5) If no feasible grasp is obtained from the first candidate model, we proceed to the next model in the priority order and repeat Steps 2-4 until a valid grasp is identified.

It should be noted that although we have original models of the environmental objects in our simulations (e.g., the platform on which the object is placed), our grasp planning does not rely on them. Instead, we can utilize the visual information in the background image to reconstruct the surrounding obstacles. This capability is validated in the final part of our experiments.

\subsection{Stability-Aware Grasp Fine-tuning}
\begin{figure}[t]
    \centering
    \includegraphics[width=0.95\linewidth]{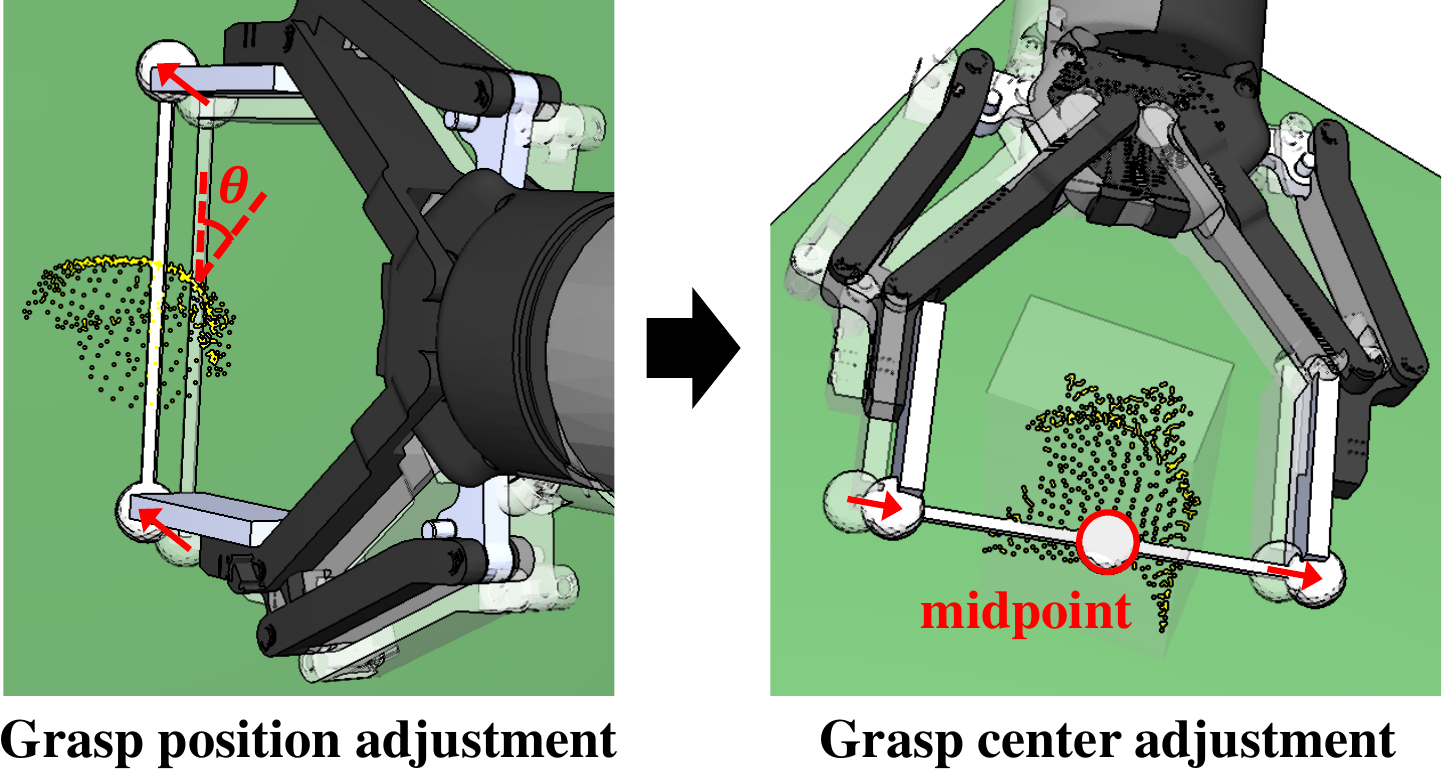}
    \caption{Two-stage grasp fine-tuning, including grasp position and grasp center adjustments to optimize the final grasp quality.}
    \label{img:7}
\end{figure}

Considering the inherent differences between similar models and real objects, imitative grasps have the potential to be unstable in real-world tasks. To evaluate and enhance their stability, we focus on the local features of contact points within the observable region of the object point cloud to implement a two-stage grasp fine-tuning process, as shown in Fig. \ref{img:7}. In the first stage, we create a stick model connecting the two ends of the gripper, denoted as $\mathbb{S}$, and use a ray-hit algorithm \cite{panda} to detect its intersections with the object point cloud $\mathbb{P}$. If no intersection is detected, the grasp is considered to be located in the unseen regions of the object and is retained as a \textit{potential grasp}, as its stability cannot be assessed. If only a single intersection point is detected, it is identified as one of the two contact points between the gripper and the object. In cases where multiple intersection points are detected, the two outermost points are regarded as the two contact points. 

For each contact point, we identify its nearest neighboring point $p_0$ in $\mathbb{P}$ and compute the acute angle $\theta$ between the normal at $p_0$ and $\mathbb{S}$. Based on this angle, we classify all feasible grasps other than \textit{potential grasps} into three types: 
\begin{enumerate}
\item{When $\theta<20^{\circ}$, the grasp is considered stable enough to be executed directly without fine-tuning; }
\item{When $\theta>40^{\circ}$, the grasp is too unstable for fine-tuning and is therefore discarded; }
\item{When $\theta\in \left [20^{\circ},40^{\circ}\right ]$, the grasp is not sufficiently stable but can be improved through fine-tuning for better quality. }
\end{enumerate}

Below, we illustrate the fine-tuning process for the case where only one contact point is detected. The two-contact-point case differs only in that $\theta$ needs to be calculated for both contact points and the criteria for grasp classification become: 1) $\forall\,\theta<20^{\circ}$; 2) $\exists\,\theta>40^{\circ}$; 3) All other cases. Additionally, when two contact points are detected, one of them is randomly selected as the reference point for fine-tuning. After fine-tuning, grasp stability is re-evaluated using the updated contact points.

In the case of $\theta\in \left [20^{\circ},40^{\circ}\right ]$, we search for the $k_1$-nearest neighbors of $p_0$ as $\left \{p_1, p_2, ..., p_{k_1} \right \}$ ordered from nearest to farthest, and then calculate $\theta$ for each neighbor starting from $p_1$ until a point satisfying $\theta<20^{\circ}$ is found. This point is marked as $p_0^*$. To prevent grasping an uneven region, we further inspect the surrounding of $p_0^*$ by searching for its $k_2$-nearest neighbors as $\left \{p_1^*, p_2^*, ..., p_{k_2}^* \right \}$. We calculate the angle between the normal of $p_0^*$ and the normal of each of its neighbor as $\theta^*$. When $\forall\,\theta^*<10^{\circ}$, the region around $p_0^*$ is considered flat and suitable for grasping. In this case, we translate the original grasp pose along the vector $\overrightarrow{p_0 p_0^*}$ to adjust the grasp position without applying any rotation. Otherwise, we continue querying the next neighboring point of $p_0$ until an eligible $p_0^*$ is found. In our task, $k_1$ should be large to capture sufficient neighborhood information, while $k_2$ should be small to focus on local geometric features. Empirically, we set $k_1=100$ and $k_2=5$ as appropriate values.

In addition to adjusting the grasp position, we recognize that the grasp stability is also influenced by the location of the grasp center. When the distances between the object and the two finger ends differ significantly, one finger end may contact the object first, potentially causing unpredictable shifting or rotating motion during the grasping process. Therefore, in the second stage, we reapply the ray-hit algorithm to detect the intersections between the updated $\mathbb{S}$ after the first-stage adjustment and the SOBB of $\mathbb{P}$. We then obtain the midpoint of the two intersection points, denoted as $p_c^*$, and translate the grasp pose along the vector $\overrightarrow{p_c p_c^*}$ to refine the grasp center, where $p_c$ denotes the original grasp center after the first-stage adjustment. In the case where two contact points are detected, their midpoint can be directly used as $p_c^*$. This grasp center refinement is also applied in the case of $\theta<20^{\circ}$.

At either fine-tuning stage, if an adjusted grasp fails to solve IK or results in a collision, we discard it and query the next grasp candidate. The retained \textit{potential grasps} are only used when all evaluable grasps have been assessed as unstable.

In practice, this fine-tuning process significantly enhances grasping performance, particularly in cases where the similarity matching results are suboptimal. In our experiments, we observe that fine-tuning is triggered in about half of the trials, boosting the grasp success rate by over 20\% compared to the baseline without fine-tuning, as demonstrated in Section \ref{ablation}.

\section{Experiments}
\subsection{Experimental Setup}
To validate our proposed method, we conduct several experiments in similarity matching and novel object grasping using a UR5e robot arm equipped with a Robotiq 2F-140 adaptive gripper and a hand-mounted RealSense D435 depth camera. To verify the generalizability of our approach under low data volume conditions, we generate a database containing no more than 100 object models derived from the YCB dataset \cite{Calli}. For each model, we preplan about 200 antipodal grasps within the gripper's width range and pre-compute the C-FPFH descriptors along with the bounding extents required for matching.

The experimental objects are placed on a fixed platform within the robot's reachable workspace, and are observed by the hand camera from a diagonal downward viewpoint. All computations are performed on a computer equipped with a Ryzen 7 5800H CPU and a GeForce RTX 3060 GPU.

To clearly showcase performance, we compare our method with several baselines, including SOTA benchmarks and the previous similarity approach. For all learning-based methods, we use their pre-trained models without additional fine-tuning.

\begin{figure}[t]
    \centering
    \includegraphics[width=\linewidth]{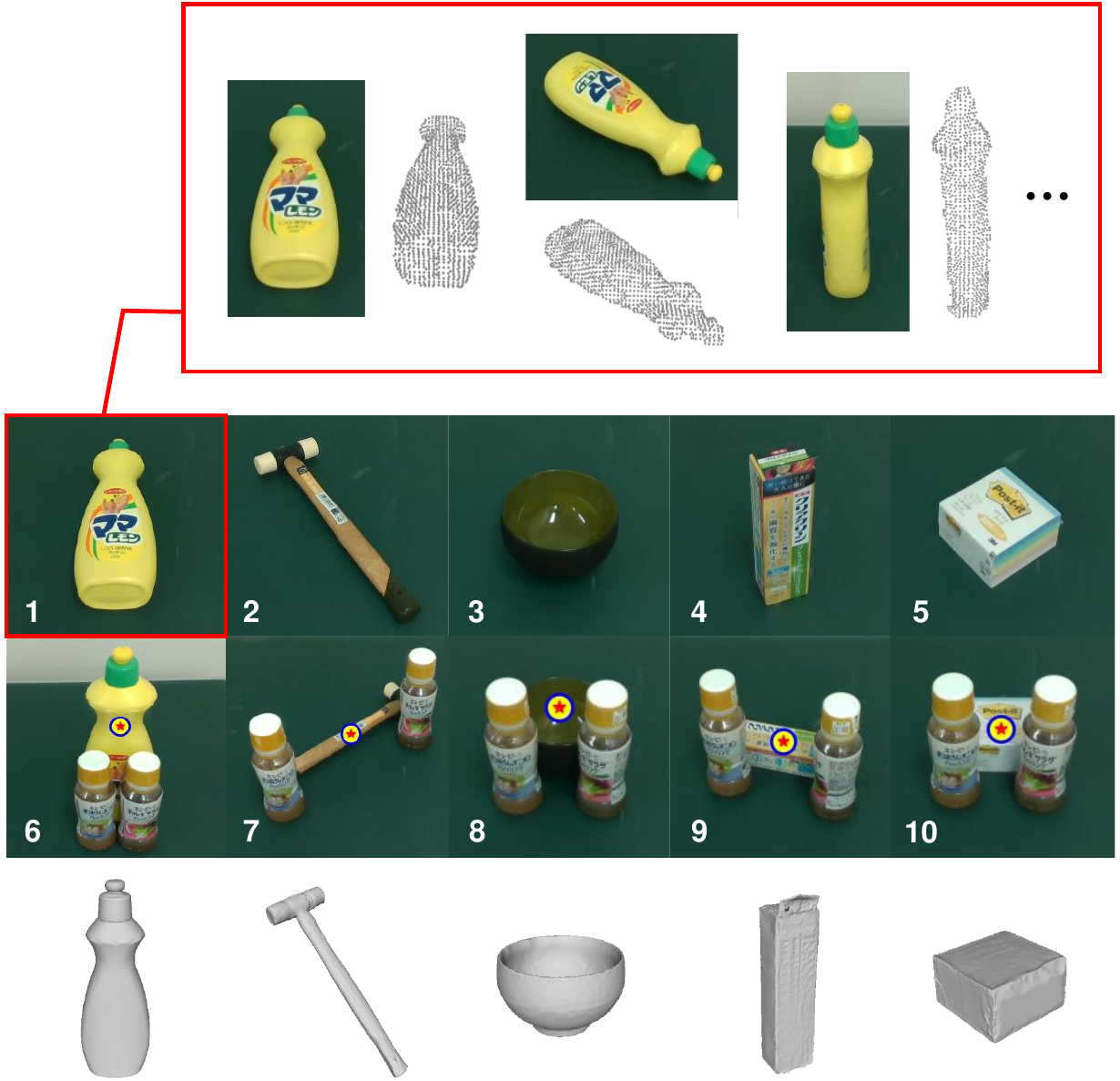}
    \caption{The experimental objects used for similarity matching. The first and second rows display the non-occluded and occluded scenes, respectively. The numbers in the figure correspond to the scene IDs in Table \ref{tab:1} and Table \ref{tab:2}. The third row presents the original 3D model of each object.}
    \label{img:8}
\end{figure}

\begin{table*}[t]
\small
\renewcommand\arraystretch{1.7}
\setlength\tabcolsep{6pt}
\centering
\caption{Evaluation Results of Similarity Matching Baselines in Non-Occluded Scenes}
\begin{tabular}{c | c c | c c | c c | c c | c c | c c}
\toprule 
\multirow{2}{*}{Scene ID} & \multicolumn{2}{c|}{1} & \multicolumn{2}{c|}{2} & \multicolumn{2}{c|}{3} & \multicolumn{2}{c|}{4} & \multicolumn{2}{c|}{5} & \multicolumn{2}{c}{Average} \\
& MA$_2$ & MT & MA$_2$ & MT & MA$_2$ & MT & MA$_2$ & MT & MA$_2$ & MT & MA$_2$ & MT \\ 
\hline 
w/o MM & 3/5 & 3.62s & 3/5 & 3.40s & 4/5 & 3.28s & 2/5 & 3.14s & 1/5 & 2.86s & 13/25 (52\%) & 3.26s \\

w/o SM & 3/5 & 0.45s & 3/5 & 0.44s & 3/5 & 0.65s & 4/5 & 0.69s & 2/5 & 0.76s & 15/25 (60\%) & 0.60s \\

w/o GM & 5/5 & 0.20s & 4/5 & 0.35s & 4/5 & 0.42s & 5/5 & 0.32s & 3/5 & 0.28s & 21/25 (84\%) & \textbf{0.31s} \\
 
w/o DM & 4/5 & 0.75s & 3/5 & 0.33s & 4/5 & 0.77s & 3/5 & 0.33s & 2/5 & 0.27s & 17/25 (64\%) & 0.49s \\

Full Matching & 5/5 & 0.30s & 4/5 & 0.35s & 4/5 & 0.58s & 5/5 & 0.30s & 4/5 & 0.21s & \textbf{22/25 (88\%)} & 0.35s \\
\bottomrule
\end{tabular}
\label{tab:1}
\end{table*}

\begin{table*}[t]
\small
\renewcommand\arraystretch{1.7}
\setlength\tabcolsep{6pt}
\centering
\caption{Evaluation Results of Similarity Matching Baselines in Occluded Scenes}
\begin{tabular}{c | c c | c c | c c | c c | c c | c c}
\toprule 
\multirow{2}{*}{Scene ID} & \multicolumn{2}{c|}{6} & \multicolumn{2}{c|}{7} & \multicolumn{2}{c|}{8} & \multicolumn{2}{c|}{9} & \multicolumn{2}{c|}{10} & \multicolumn{2}{c}{Average} \\
& MA$_5$ & MT & MA$_5$ & MT & MA$_5$ & MT & MA$_5$ & MT & MA$_5$ & MT & MA$_5$ & MT \\ 
\hline 
w/o MM & 2/5 & 3.29s & 1/5 & 2.92s & 1/5 & 2.58s & 2/5 & 2.83s & 1/5 & 2.06s & 7/25 (28\%) & 2.74s \\

w/o SM & 3/5 & 1.21s & 2/5 & 0.97s & 3/5 & 0.72s & 3/5 & 1.14s & 4/5 & 0.72s & 15/25 (60\%) & 0.95s \\

w/o GM & 3/5 & 1.67s & 2/5 & 0.26s & 2/5 & 0.45s & 3/5 & 0.83s & 2/5 & 0.63s & 12/25 (48\%) & 0.77s \\
 
w/o DM & 4/5 & 0.48s & 4/5 & 0.39s & 4/5 & 0.49s & 3/5 & 0.52s & 4/5 & 0.43s & \textbf{19/25 (76\%)} & 0.46s \\

Full Matching & 4/5 & 0.41s & 3/5 & 0.21s & 3/5 & 0.38s & 4/5 & 0.34s & 4/5 & 0.36s & 18/25 (72\%) & \textbf{0.34s} \\
\bottomrule
\end{tabular}
\label{tab:2}
\end{table*}

\subsection{Evaluation of Similarity Matching}
To verify the effectiveness of our proposed similarity matching approach, we carry out comparative experiments using the following baselines: 1) \textbf{w/o MM} (Without Multi-level Matching); 2) \textbf{w/o SM} (Without Semantic Matching); 3) \textbf{w/o GM} (Without Geometric Matching); 4) \textbf{w/o DM} (Without Dimensional Matching); 5) \textbf{Full Matching}. For the first baseline, we directly apply point cloud registration to rank all database models from most to least similar without performing multi-level matching. In the second through fourth baselines, we perform incomplete matching by excluding one of the three levels, followed by the point cloud registration process to rank candidates. The last baseline uses the complete method.

\begin{figure*}[t]
    \centering
    \includegraphics[width=0.95\linewidth]{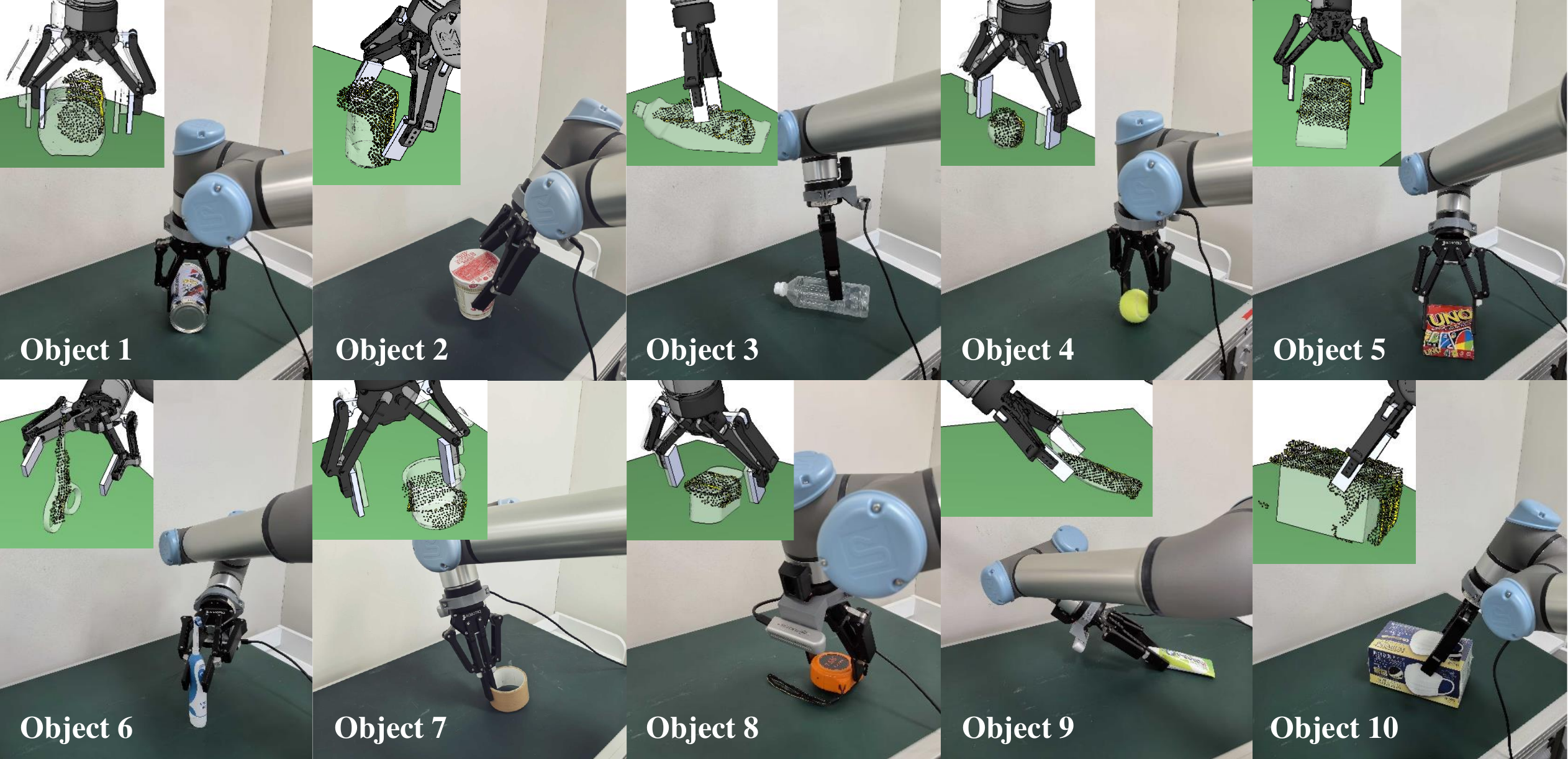}
    \caption{Grasping various novel objects using our similarity-based method. The top left corner of each figure displays the single-view object point cloud, the matched candidate model, and the grasp planning with fine-tuning process. The object IDs shown in the bottom-left correspond to those in Table \ref{tab:3}.}
    \label{img:9}
\end{figure*}

To accurately assess the performance of similarity evaluation across different baselines, we utilize 5 different types of experimental objects with their original 3D models, as shown in Fig. \ref{img:8}, and incorporate them into the database. During the matching process, if the original model of the target object is identified within the highest priority list of candidate models, we consider it an accurate match. Based on this principle, we develop the following two evaluation metrics for the matching results: 1) \textbf{Matching Accuracy (MA)}, which is defined as: 
\begin{align*}
    {\rm MA}_x = \frac{{\rm Number\;of\;accurate\;matches}}{{\rm Total\;number\;of\;matching\;attempts}}
\end{align*}
Here, $x$ indicates that a match is considered accurate if the original model appears in the top $x$ candidates. We need this $x$ to be adaptable to different detection conditions and to allow for matching errors when the database contains existing models similar to the newly added objects. 2) \textbf{Matching Time (MT)}, which records the total time taken for the matching process, measured in seconds (s), from the start of similarity matching to the completion of point cloud registration.

We first test in non-occluded scenes, where the objects are placed on a stationary platform without surrounding obstacles. For each object, we obtain its visual features from 5 different observation angles and use these features to perform each baseline, generating the corresponding matching results. We set $x=2$ as a strict criterion in this case to clearly distinguish the performance of different baselines. From the experimental results shown in Table \ref{tab:1}, the comparison between \textbf{w/o MM} and \textbf{Full Matching} highlights the advantage of using multi-level similarity matching for achieving significantly higher matching accuracy while reducing computation time. The reason for this discrepancy is that without multi-level matching, we need to perform point cloud registration with all database models, the results of which are susceptible to the uncertainty of partial observation, resulting in low accuracy and efficiency. Besides, the absence of either semantic or dimensional levels in matching leads to a noticeable performance drop compared to the full matching method, emphasizing their importance in effectively narrowing down the range of similar candidates. On the other hand, however, the baseline \textbf{w/o GM} obtains very similar results to \textbf{Full Matching}, failing to demonstrate the effectiveness of our geometric matching method. This occurs because the database contains the categories of all experimental objects, making semantic and dimensional matching alone sufficient for achieving accurate results.

Therefore, for further validation, we test in occluded scenes where the objects are partially occluded by two small bottles acting as obstacles. During visual detection, we manually place a marker in the image (see Fig. \ref{img:8}) to extract only the recognition result of the target object. Similar to the non-occluded scenes, we perform 5 matching attempts for each object and baseline using various observation angles and occlusion conditions (in principle, occluding no more than half of the object to ensure matchability). The difference is that in this case we set $x=5$ to allow for larger matching errors due to the increased visual uncertainty. From the experimental results shown in Table \ref{tab:2}, we observe that the accuracy of \textbf{w/o MM} becomes worse due to the sparser visual information, whereas the performance of \textbf{Full Matching} remains relatively stable. More importantly, the advantage of using C-FPFH-based geometric matching is clearly demonstrated by a significant performance drop in the baseline \textbf{w/o GM}. The reason for this is that in occluded scenes, the reliability of both semantic and dimensional matching declines due to incomplete object recognition, making the matching accuracy heavily dependent on the results of geometric matching. We even find that in some cases, dimensional matching negatively impacts the matching results, whereas semantic matching still helps to achieve accurate results when the occluded object is correctly recognized. Additionally, the full matching method achieves the lowest computation time by narrowing down similar candidates to a smaller range, while other baselines yield more candidates when a certain matching level is absent, leading to longer computation times.

By combining the results from both scenes, our complete similarity matching method demonstrates optimal performance in terms of accuracy and efficiency under varying detection conditions and across different object types.

\subsection{Grasping Isolated Objects}\label{single}
To verify the performance of novel object grasping using our method, we conduct grasping experiments in two scenarios: isolated and cluttered scenes. We first focus on single-object grasping by comparing our proposed method with two learning-based benchmarks, \textbf{PointNetGPD} \cite{Liang} and \textbf{3DSGrasp} \cite{Mohammadi}, as well as the \textbf{previous similarity approach} \cite{Chen}. All these approaches address the grasping of isolated novel objects by leveraging object point clouds. The key differences are that, PointNetGPD directly predicts grasps from single-view point clouds; 3DSGrasp performs shape completion to reconstruct unseen regions of the point cloud and plans grasps based on the refined cloud; the previous similarity approach utilizes multi-view point clouds for similarity matching, followed by grasp knowledge transfer from similar references to the unknown target. For PointNetGPD and 3DSGrasp, we rank the grasp candidates based on their quality scores and proceed sequentially from the highest-scoring grasp until a feasible one is found. For both the previous and our similarity approaches, we process the grasp candidates in no specific~order and continue the computation until a feasible grasp is identified.

We use three evaluation metrics to assess the performance of each method: 1) \textbf{Grasp Success Rate (GSR)}; 2) \textbf{Plan Success Rate (PSR)}; 3) \textbf{Average Planning Time (APT)}. Considering three possible outcomes in a grasping task: I) Grasp planning fails (no grasp output); II) Grasp planning succeeds, but the output grasp fails to catch or lift the object; III) Grasp planning succeeds, and the output grasp successfully catches and lifts the object, we define GSR and PSR as follows:
\begin{align*}
    {\rm GSR} = \frac{{\rm Num(III)}}{{\rm Num(II+III)}},\;
    {\rm PSR} = \frac{{\rm Num(II+III)}}{{\rm Num(I+II+III)}}
\end{align*}
where $\rm Num()$ denotes the total number of outcomes contained in the bracket. APT is the approximate average of the duration from the start of similarity matching to the completion of grasp planning, recorded only when the grasp planning is successful. These three metrics represent the accuracy, generalizability, and efficiency of each implemented method, respectively.

\begin{table*}[t]
\small
\renewcommand\arraystretch{1.7}
\setlength\tabcolsep{6pt}
\centering
\caption{Experimental Results of Grasping Isolated Objects Compared with Benchmarks}
\begin{tabular}{c | c | c c c c c c c c c c | c c}
\toprule 
\multicolumn{2}{c|}{Object ID} & 1 & 2 & 3 & 4 & 5 & 6 & 7 & 8 & 9 & 10 & Average & APT \\
\hline 
\multirow{2}{*}{PointNetGPD} & GSR & 7/8 & 6/8 & 4/6 & 2/4 & - & 3/6 & 8/9 & 4/7 & 3/4 & 2/6 & 39/58 (67\%) & \multirow{2}{*}{$\approx$ 6s} \\
& PSR & 8/10 & 8/10 & 6/10 & 4/10 & 0/10 & 6/10 & 9/10 & 7/10 & 4/10 & 6/10 & 58/100 (58\%) \\
\hline 
\multirow{2}{*}{3DSGrasp} & GSR & 7/10 & 3/10 & 5/7 & - & 2/2 & 2/4 & 8/8 & 0/6 & 3/4 & 6/10 & 36/61 (59\%) & \multirow{2}{*}{$\approx$ 8s} \\
& PSR & 10/10 & 10/10 & 7/10 & 0/10 & 2/10 & 4/10 & 8/10 & 6/10 & 4/10 & 10/10 & 61/100 (61\%) \\
\hline 
\multirow{2}{*}{\makecell{Previous simi-\\larity approach}} & GSR & 8/10 & 7/10 & 4/7 & 10/10 & 7/8 & 10/10 & 9/10 & 5/8 & 5/6 & 4/10 & 69/89 (78\%) & \multirow{2}{*}{$\approx$ 5s} \\
& PSR & 10/10 & 10/10 & 7/10 & 10/10 & 8/10 & 10/10 & 10/10 & 8/10 & 6/10 & 10/10 & 89/100 (89\%) \\
\hline 
\multirow{2}{*}{Our method} & GSR & 10/10 & 10/10 & 8/9 & 9/10 & 9/9 & 10/10 & 10/10 & 8/10 & 10/10 & 10/10 & \textbf{94/98 (96\%)} & \textbf{\multirow{2}{*}{$\boldsymbol\approx$ 2s}} \\
& PSR & 10/10 & 10/10 & 9/10 & 10/10 & 9/10 & 10/10 & 10/10 & 10/10 & 10/10 & 10/10 & \textbf{98/100 (98\%)} \\
\bottomrule
\end{tabular}
\label{tab:3}
\end{table*}

\begin{figure*}[t]
    \centering
    \includegraphics[width=\linewidth]{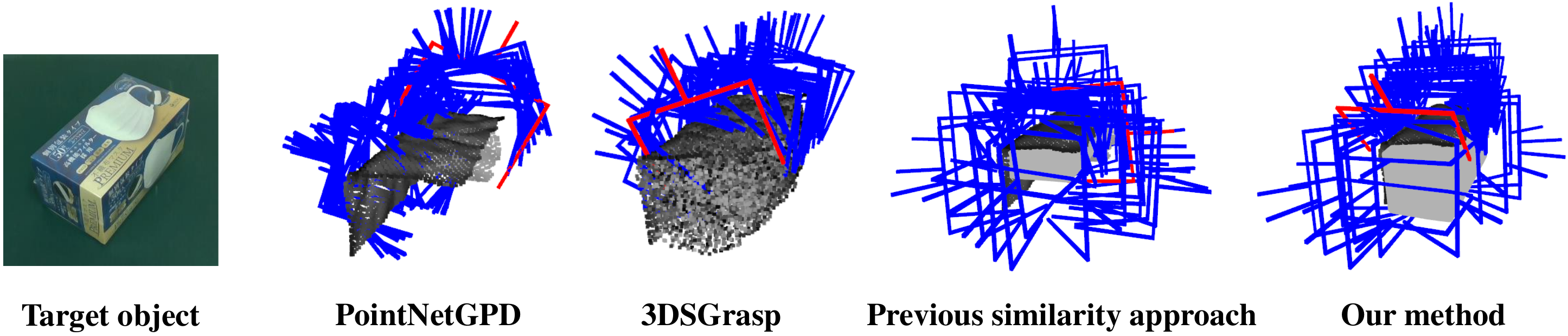}
    \caption{Comparison of different grasp planning methods based on object point clouds. The blue grasps represent all the generated grasp candidates, while the red grasps indicate the final executed grasps that are first recognized as IK-solvable and collision-free from the candidate list.}
    \label{img:10}
\end{figure*}

For the experimental setup, we select 10 previously unseen objects varying in category, shape and size (see Fig.~\ref{img:9}), and place them in arbitrary poses on a stationary platform within the camera's field of view and the robot's reachable workspace. During each trial, we apply our object recognition method to obtain the single-view point cloud of the target object and execute the corresponding grasp planning baseline based on this point cloud. For each object and method, we perform 10 planning and grasping attempts, considering a grasp successful if the object is steadily caught and lifted to a specified height (approximately 20 cm). As can be seen from the results shown in Table \ref{tab:3}, our proposed method significantly outperforms all other methods in terms of both success rate and computational efficiency, demonstrating its superior overall performance.

For the learning-based methods, we observe that both PointNetGPD and 3DSGrasp achieve very low PSRs for specific objects, such as Objects 4 and 5. Although their network outputs provide reasonable grasp poses, most of these poses are inexecutable (IK unsolved or collisions detected) due to improper grasp positions and orientations, which is likely to be an inherent limitation of their training models. The previous similarity approach can successfully generate executable grasps in most cases; however, its GSR remains unsatisfactory, primarily due to the gap between multi-view and single-view detection conditions, the limitations of score-based matching, and the absence of subsequent fine-tuning.

\begin{figure}[t]
    \centering
    \includegraphics[width=0.95\linewidth]{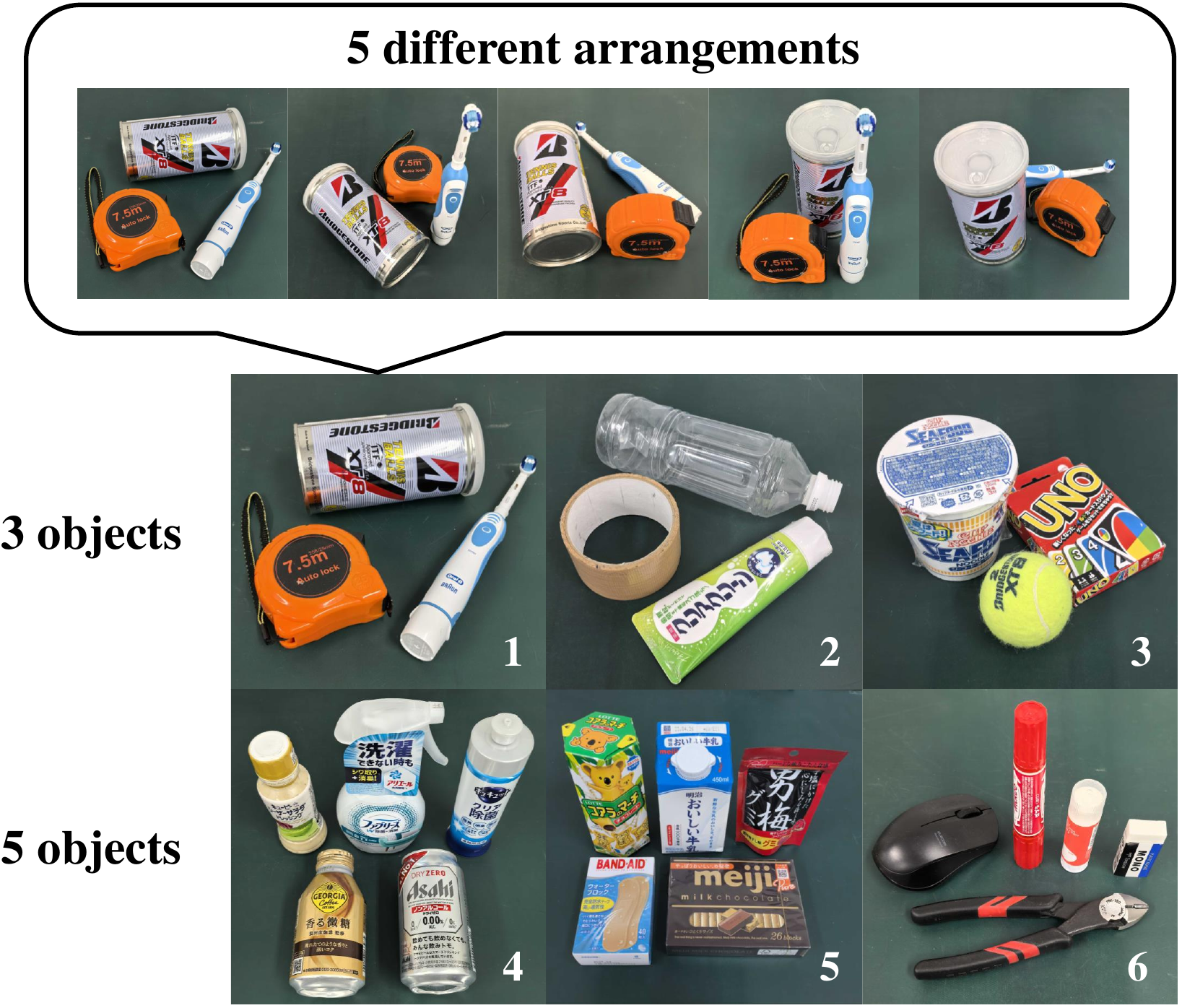}
    \caption{Objects in clutter used for grasping experiments. Each clutter set contains 3 or 5 objects, and is tested in 5 different arrangements.}
    \label{img:11}
\end{figure}

An example visualization of the grasp planning results using different methods is shown in Fig. \ref{img:10}. The grasp candidates generated by PointNetGPD are mostly unstable, such as grasping the object edge from an improper angle. 3DSGrasp can refine the partial point cloud to approximate the actual object shape; however, its errors in point refinement can still lead to suboptimal grasps that lack sufficient stability. This limitation is specifically noticeable for objects with uneven and smooth surfaces, such as Objects 2 and 8. The previous similarity approach can generate more reasonable and diverse grasps by utilizing a similar reference model; however, an inappropriate matching result derived from scoring functions may still lead to infeasible grasps, such as those colliding with the unseen regions of the target object. Similar issues also arise when basic antipodal grasp sampling is applied to a single-view object point cloud, where the unseen regions introduce uncertainty in both grasp robustness and collision risk, often resulting in failure. Additionally, its grasp stability cannot be guaranteed due to the lack of a fine-tuning process. In contrast, our method achieves optimal grasp planning by leveraging an accurately matched reference model and improving the grasp quality through a two-stage fine-tuning approach.

Regarding APT, the learning-based methods require additional time for grasp sampling or point cloud completion, leading to longer computation durations. The previous similarity approach requires matching with all models in the database, making it also time-intensive. In contrast, our method pre-sorts candidate models using a multi-level similarity matching approach, cutting computational time by more than half compared to other methods.

\subsection{Grasping Cluttered Objects}

\begin{table*}[t]
\small
\renewcommand\arraystretch{1.7}
\setlength\tabcolsep{7pt}
\centering
\caption{Experimental Results of Grasping Cluttered Objects Compared with Benchmarks}
\begin{tabular}{c | c | c c c | c | c c c | c}
\toprule 
\multicolumn{2}{c|}{\multirow{2}{*}{Clutter ID}} & \multicolumn{3}{c|}{3 objects} & \multirow{2}{*}{Average} & \multicolumn{3}{c|}{5 objects} & \multirow{2}{*}{Average} \\
\multicolumn{2}{c|}{} & 1 & 2 & 3 & & 4 & 5 & 6 \\
\hline 
\multirow{2}{*}{GraspNet} & GSR & 7/13 & 8/13 & 8/12 & 23/38 (60.5\%) & 12/16 & 8/11 & - & 20/27 (74.1\%) \\
& DR & 7/15 & 8/15 & 8/15 & 23/45 (51.1\%) & 12/15 & 8/15 & 0/15 & 20/45 (44.4\%) \\
\hline 
\multirow{2}{*}{HGGD} & GSR & 9/17 & 6/10 & 10/12 & 25/39 (64.1\%) & 13/17 & 9/13 & 2/2 & 24/32 (75.0\%) \\
& DR & 9/15 & 6/15 & 10/15 & 25/45 (55.6\%) & 13/15 & 9/15 & 2/15 & 24/45 (53.3\%) \\
\hline 
\multirow{2}{*}{Our method} & GSR & 15/16 & 14/16 & 14/14 & \textbf{43/46 (93.5\%)} & 15/17 & 14/16 & 13/14 & \textbf{42/47 (89.4\%)} \\
& DR & 15/15 & 14/15 & 14/15 & \textbf{43/45 (95.6\%)} & 15/15 & 14/15 & 13/15 & \textbf{42/45 (93.3\%)} \\
\bottomrule
\end{tabular}
\label{tab:4}
\end{table*}

\begin{figure*}[t]
    \centering
    \includegraphics[width=\linewidth]{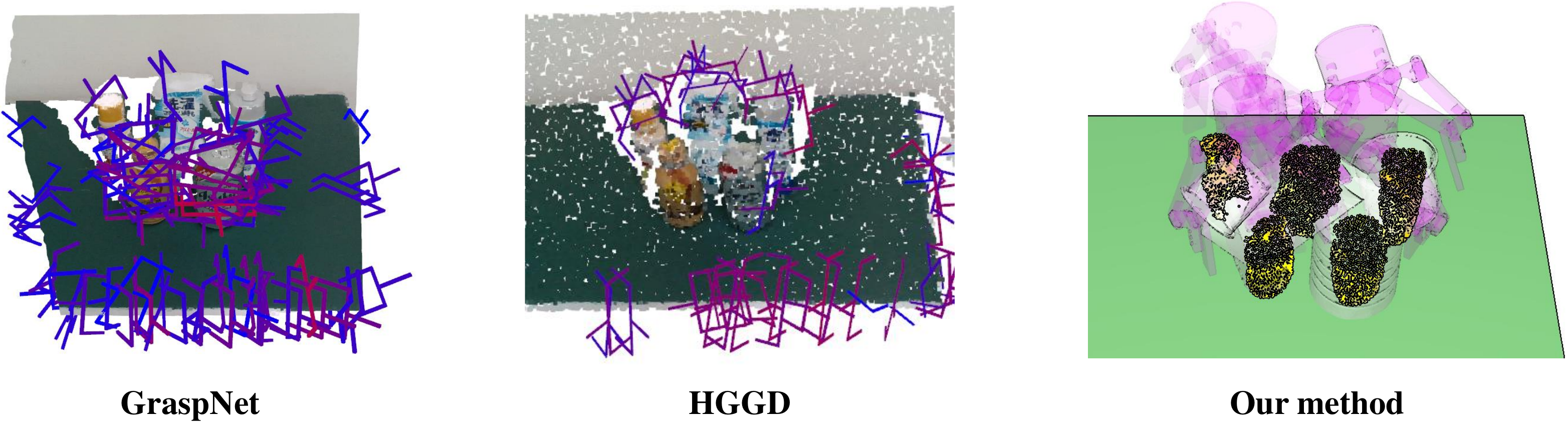}
    \caption{Comparison of our method with grasping benchmarks for objects in clutter. In the outputs of GraspNet and HGGD, red grasps indicate high-quality grasps, while blue grasps indicate low-quality ones. Our method generates high-quality grasps for all target objects by rendering similar reference models.}
    \label{img:12}
\end{figure*}

For further validation of object grasping performance, we conduct experiments in cluttered scenes, where multiple objects are randomly arranged to create more challenging grasping scenarios. We compare our approach with two SOTA methods: a widely known large-scale benchmark for general object grasping, \textbf{GraspNet} \cite{Fang}, and a recent advanced work on grasping objects in clutter, \textbf{HGGD} \cite{Siang}. Both methods detect grasps directly from the scene RGBD input, making it difficult to distinguish which grasp corresponds to which object. Therefore, for these two methods, we develop the following steps to achieve the task of clearing cluttered objects: 1) Capture the RGBD image of the scene and apply the grasp detection method; 2) Rank all detected grasps based on their evaluated quality scores; 3) Compute IK iteratively until the first executable grasp in the ranked list is found; 4) Execute the identified grasp to remove the corresponding object from the scene, return the robot to its initial pose, and repeat the above process. In contrast, our method completes detection and computation at once by simultaneously performing similarity matching and grasp planning for all objects in the clutter. We achieve object recognition similarly to the single-object scene, but with the difference that all recognition results (excluding redundant ones) within the clutter region are extracted and processed in separate threads. During grasp planning, the following principles are applied: I) When planning grasps for one object, all other objects are treated as obstacles; II) Once a feasible grasp is generated for an object, it is removed from the obstacle list to facilitate planning for the remaining objects. We also record the grasp generation sequence and execute the planned grasps in the corresponding order.

To assess task performance, we utilize two evaluation metrics: 1) \textbf{Grasp Success Rate (GSR)}; 2) \textbf{Declutter Rate (DR)}. GSR is computed similarly to the single-object scene, but differs in that a single re-planning attempt is allowed after a grasp failure, in which case the number of executed grasps may exceed the total number of objects in the clutter. DR evaluates the task completion rate and is calculated as:
\begin{align*}
    {\rm DR} = \frac{{\rm Number\;of\;successfully\;removed\;objects}}{{\rm Total\;number\;of\;objects\;to\;be\;removed}}
\end{align*}

For the experimental setup, we leverage two types of clutter configurations as shown in Fig. \ref{img:11}: Clutters 1-3 consist of 3 objects selected from the single-object grasping experiment, while Clutters 4-6 comprise 5 objects with a broader variety of types. We perform 5 attempts for each clutter set under different object arrangements, each featuring varying degrees of occlusion. To avoid intermediate collisions unrelated to the grasp pose, we design a pregrasp pose by retracting 8 cm from the grasp pose and a leaving pose by moving 20 cm vertically upwards from the grasp pose. The robot follows this sequence of movements: initial pose $\to$ pregrasp pose $\to$ grasp pose (gripper closed) $\to$ leaving pose $\to$ placement pose (gripper open) $\to$ initial pose. A task stops midway when there is no executable grasp output or when two consecutive grasp failures occur. From the experimental results shown in Table \ref{tab:4}, we observe that GraspNet and HGGD exhibit similar performance in both GSR and DR due to their respective advantages of large-scale training and well-designed learning frameworks. However, they struggle to handle specific objects, such as the small and thin items in Clutter 6, where we observe that very few viable grasps can be generated. Such failures can be attributed to the inherent limitations of learning-based approaches when the sensing condition and operating environment differ significantly from their training setup. 

Additionally, as shown in Fig.~\ref{img:12}, GraspNet fails to generate high-quality grasps for occluded objects positioned at the back. Meanwhile, HGGD can generate a few good grasps for the objects behind, but they are too sparse to ensure an executable grasp, and the front objects are ignored in this case. In contrast, our method efficiently generates robust grasps for both non-occluded and occluded objects. Even when the matching results are suboptimal due to sparse and noisy visual inputs, the subsequent fine-tuning process in our method ensures high final grasp quality. Therefore, failures in grasp planning and execution are rarely observed with our approach, demonstrating its superior performance across all types of clutter sets and object arrangements.

\subsection{Ablation Study}\label{ablation}
To validate the effectiveness of the main components in our approach, we conduct ablation studies using the same baselines from the similarity matching experiment, including \textbf{w/o MM}, \textbf{w/o SM}, \textbf{w/o GM}, and \textbf{w/o DM}. These baselines exclude one or all three levels in similarity matching, relying primarily on the processes of point cloud registration and grasp fine-tuning to generate final grasps. In addition, we introduce an extra baseline, \textbf{w/o GF} (Without Grasp Fine-tuning), which applies the complete matching method but excludes the two-stage fine-tuning process during grasp planning.

\begin{figure}[t]
    \centering
    \includegraphics[width=\linewidth]{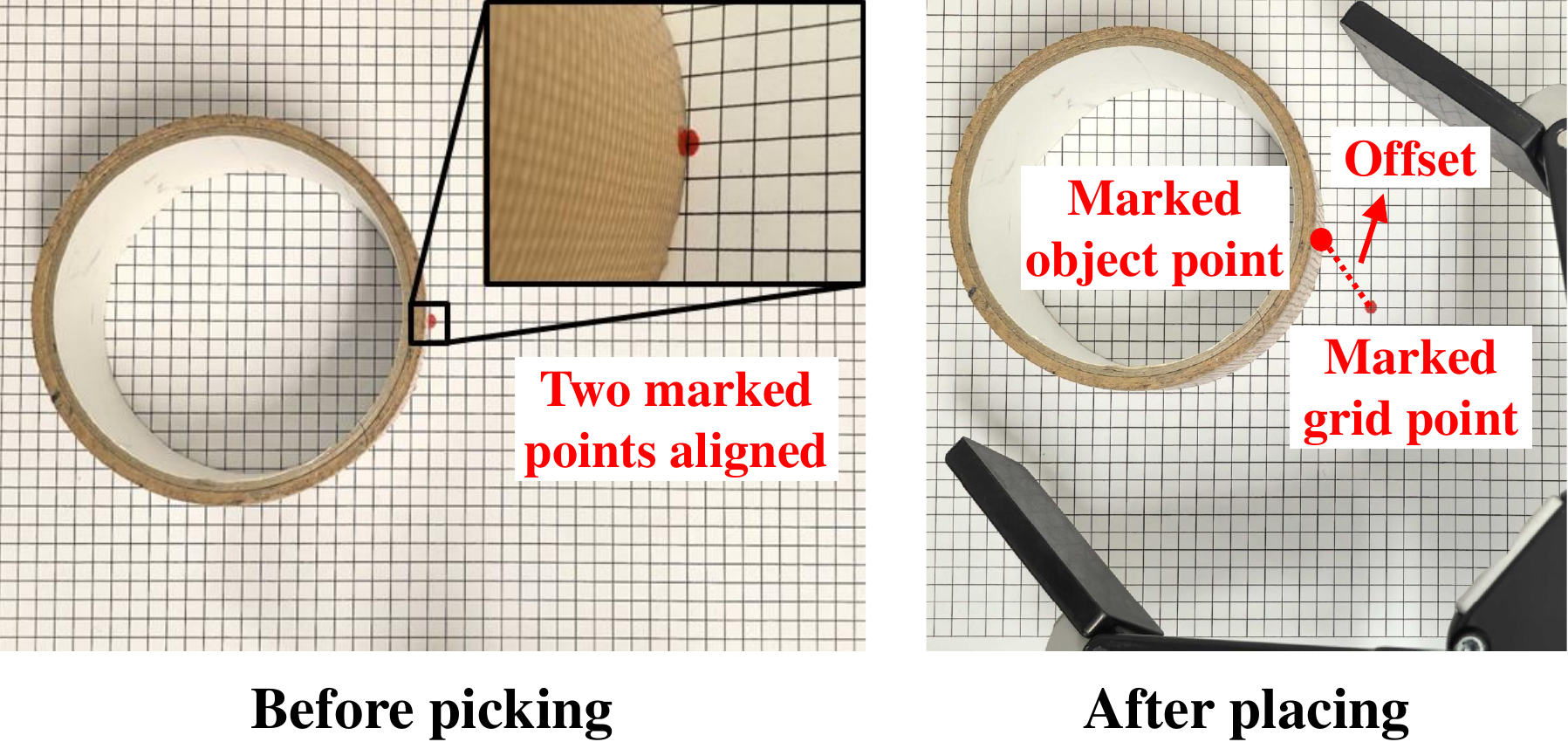}
    \caption{Evaluation of grasp quality using a graph paper. After a pick-and-place motion, the stability of the executed grasp is quantified by measuring the offset between a marker on the object and an initial point on the paper.}
    \label{img:13}
\end{figure}

We reuse the objects from the single-object grasping experiment to conduct trials. To better distinguish the performance of different baselines, in addition to \textbf{Grasp Success Rate (GSR)}, we introduce a new evaluation metric called \textbf{Average Offset (AO)}. This metric is measured using a graph paper with 5 mm grid units. At the initial stage, we mark a specific grid point on the paper and a corresponding point on the edge of the target object, aligning them perfectly (see Fig. \ref{img:13}). After planning a grasp, the robot executes the grasp, lifts the object, returns to the initial pose, and places the object back in its original position. During this process, the object may remain stationary or shift slightly within the gripper, depending on the stability of the grasp. Once the object is placed back, we locate the marked point on the object, identify the nearest grid point, and calculate its distance from the original marked grid point as the offset value. A smaller offset value indicates higher grasp quality. The use of graph paper offers a significant advantage, allowing efficient distance measurement based on the grid size without requiring complex procedures, as high precision is unnecessary. The offset is disregarded in cases of grasp failure. For each object and baseline, we perform 5 attempts and compute the average of the resulting offsets (AO) in both non-occluded and occluded scenes. In the occluded scenes, similar to the similarity matching experiment, two small bottles are used as obstacles during visual detection but are removed before grasp execution.

\begin{table}[t]
\small
\renewcommand\arraystretch{1.7}
\setlength\tabcolsep{7pt}
\centering
\caption{Ablation Studies On Each Component of Our Method}
\begin{tabular}{c | c c | c c}
\toprule 
& \multicolumn{2}{c|}{Non-occluded scene} & \multicolumn{2}{c}{Occluded scene} \\
& GSR $\uparrow$ & AO $\downarrow$ & GSR $\uparrow$ & AO $\downarrow$ \\
\hline 
w/o MM & 74\% & 11.4 mm & 54\% & 25.3 mm \\
w/o SM & 82\% & 10.5 mm & 74\% & 17.0 mm \\
w/o GM & 90\% & \textbf{6.1 mm} & 66\% & 21.4 mm \\
w/o DM & 78\% & 9.4 mm & 72\% & 15.5 mm \\
w/o GF & 72\% & 15.2 mm & 56\% & 24.1 mm \\
Full method & \textbf{92\%} & 7.1 mm & \textbf{84\%} & \textbf{11.1 mm} \\
\bottomrule
\end{tabular}
\label{tab:5}
\end{table}

\begin{figure*}[t]
    \centering
    \includegraphics[width=\linewidth]{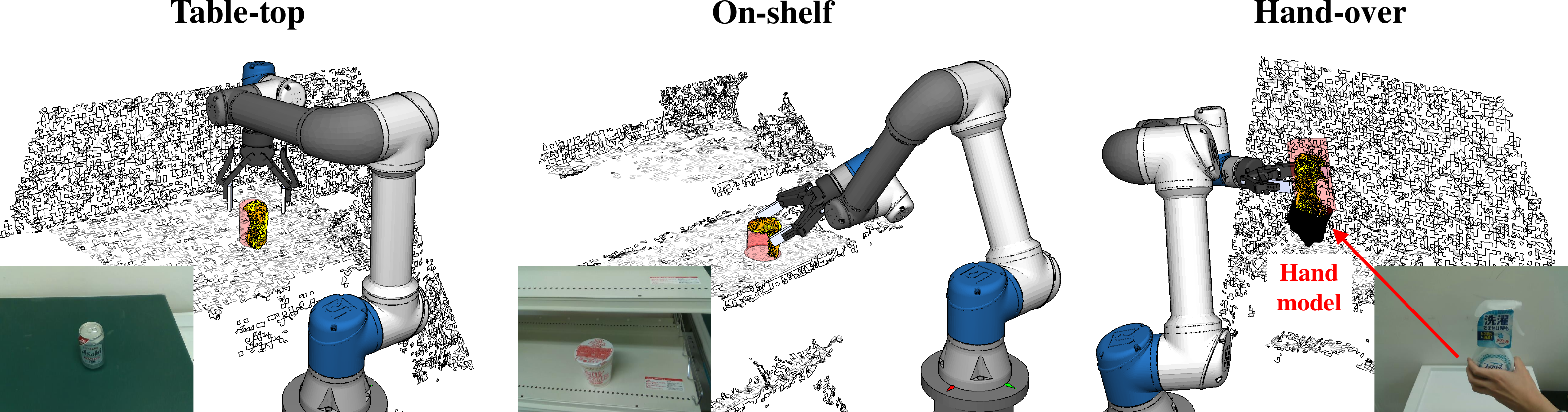}
    \caption{Application of our method to achieve grasping tasks in three different scenarios without the original models of the environmental objects.}
    \label{img:14}
\end{figure*}

\begin{figure}[t]
    \centering
    \includegraphics[width=\linewidth]{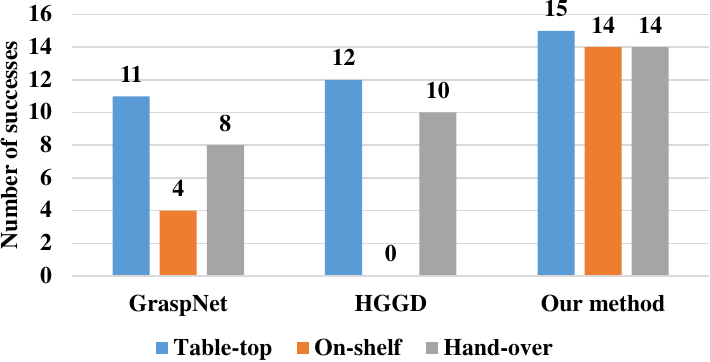}
    \caption{Comparison of the performance robustness of different methods.}
    \label{img:15}
\end{figure}

As shown in the experimental results in Table \ref{tab:5}, our full method consistently delivers optimal performance across both scenes, highlighting the effectiveness of all components in our approach. The baseline \textbf{w/o GM} achieves comparable results in the non-occluded scene but suffers a noticeable performance drop in the occluded scene, which aligns with the findings from the similarity matching experiment and confirms that our proposed C-FPFH descriptor is specifically effective in handling occlusions. Furthermore, the importance of multi-level matching and grasp fine-tuning is surprisingly close, as evidenced by the results of \textbf{w/o MM} and \textbf{w/o GF}. This underscores that both accurate matching and an effective fine-tuning process are critical for achieving high-quality grasps.

\subsection{Robustness to Environmental Changes}

\begin{figure}[t]
    \centering
    \includegraphics[width=\linewidth]{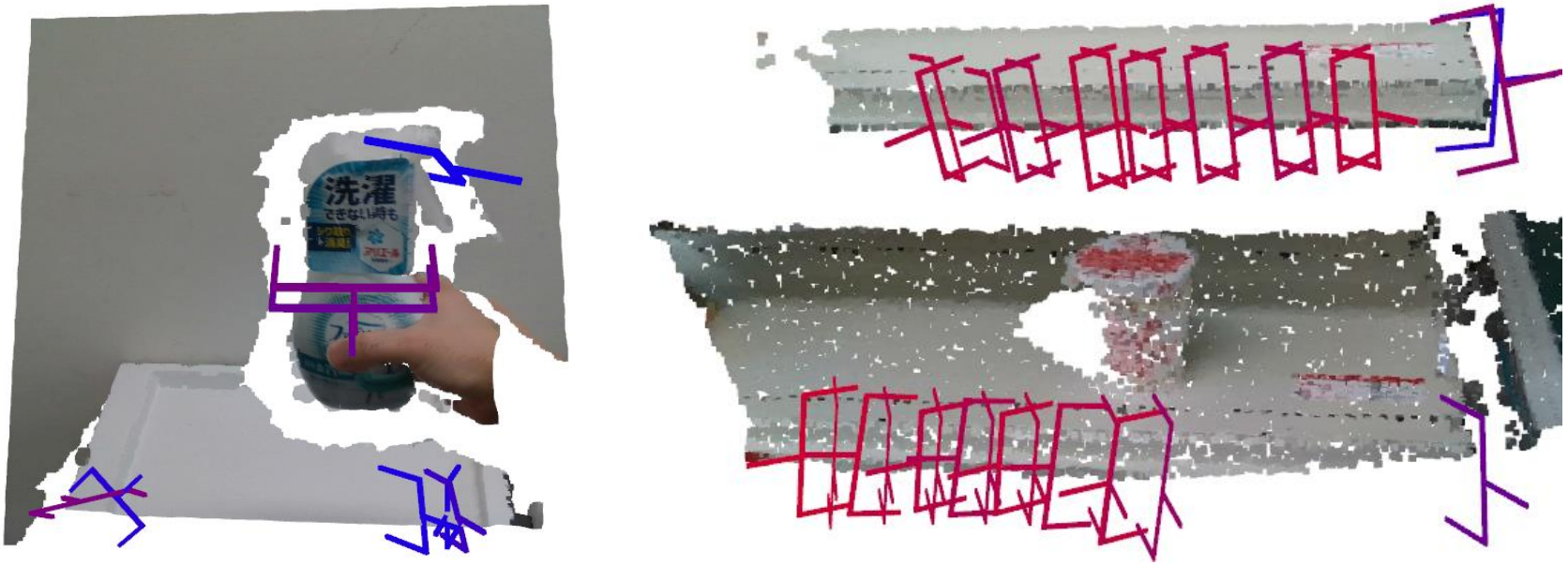}
    \caption{Failures of learning-based methods in on-shelf and hand-over tasks.}
    \label{img:16}
\end{figure}

In the final experiment, we validate the robustness of our proposed method across varying environments by performing grasping tasks in three different scenarios: \textbf{table-top}, \textbf{on-shelf}, and \textbf{hand-over}, as shown in Fig. \ref{img:14}. Unlike the previous experiments, we assume that the original models of environmental objects such as the platform are not available, requiring all environmental features to be acquired through visual detection. To handle this, we utilize the RGBD information captured from the background image to reconstruct a scene model based on the point cloud features. This reconstructed model is then used for collision detection during grasp planning, which demonstrates that our method is independent of pre-existing environmental object models. In the hand-over task, we further recover a hand model based on the recognition of the human hand to prevent grasps that may collide with the hand.

We compare the performance of our method with two learning-based benchmarks, GraspNet and HGGD, using the 3 objects shown in Fig. \ref{img:14}, for which all the methods can generate feasible grasps to ensure comparability. For each object and method, we perform 5 grasping attempts and record the number of successful attempts. The results, summarized in Fig. \ref{img:15}, clearly demonstrate that our method consistently achieves robust and high performance across all scenarios, whereas both GraspNet and HGGD exhibit significant variability in different environments, highlighting their sensitivity to environmental changes. As illustrated by the failure cases in Fig. \ref{img:16}, the learning-based methods may produce very few grasp candidates for the target object when it is positioned close to the camera during a hand-over task, and tend to detect high-quality grasps on environmental objects rather than the target object during an on-shelf task. These observations underscore the effectiveness of our method in addressing the key limitations of learning-based approaches.

\section{Conclusions}
In this study, we present a novel framework for single-view object grasping by introducing a multi-level similarity matching approach that accurately identifies similar reference models from an existing database to guide the grasping of unknown target objects. The matching process simultaneously evaluates object similarity from the aspects of semantics, geometry, and dimensions to optimize the selection of potential candidate models. Notably, we introduce the C-FPFH descriptor, a novel geometric descriptor, which efficiently evaluates the similarity between partial point clouds from observed objects and complete point clouds from database models. This descriptor demonstrates exceptional effectiveness in handling occlusions. Additionally, we integrate LLM to assist with semantic matching, propose the SOBB for accurate dimensional matching, develop a PDM-based point cloud registration method to achieve imitative grasp planning, and incorporate a two-stage grasp fine-tuning process to optimize the final grasp quality. Through extensive real-world experiments, our method significantly outperforms existing benchmarks across all metrics for grasping unknown objects in both isolated and cluttered scenes, exhibiting remarkable robustness to sensing noise and environmental changes, a capability that current learning-based approaches cannot reach.

In the future, we plan to extend this method to dynamic scenarios and more complex manipulation tasks, where additional knowledge can be integrated beyond simple grasping.

\appendices
\section{}\label{details}
In this appendix, we present detailed implementation insights into the core components of our approach, including how we leverage the mutually compensatory nature of multi-level similarity matching to select suitable candidates, and the rationale for using the two principal components of FPFH descriptors in constructing the C-FPFH descriptors. Additionally, we conduct a case study to demonstrate the robustness of our method under varying levels of sensing noise.

\subsection{Potential Mechanisms for Candidate Selection}
\begin{figure}[t]
    \centering
    \includegraphics[width=\linewidth]{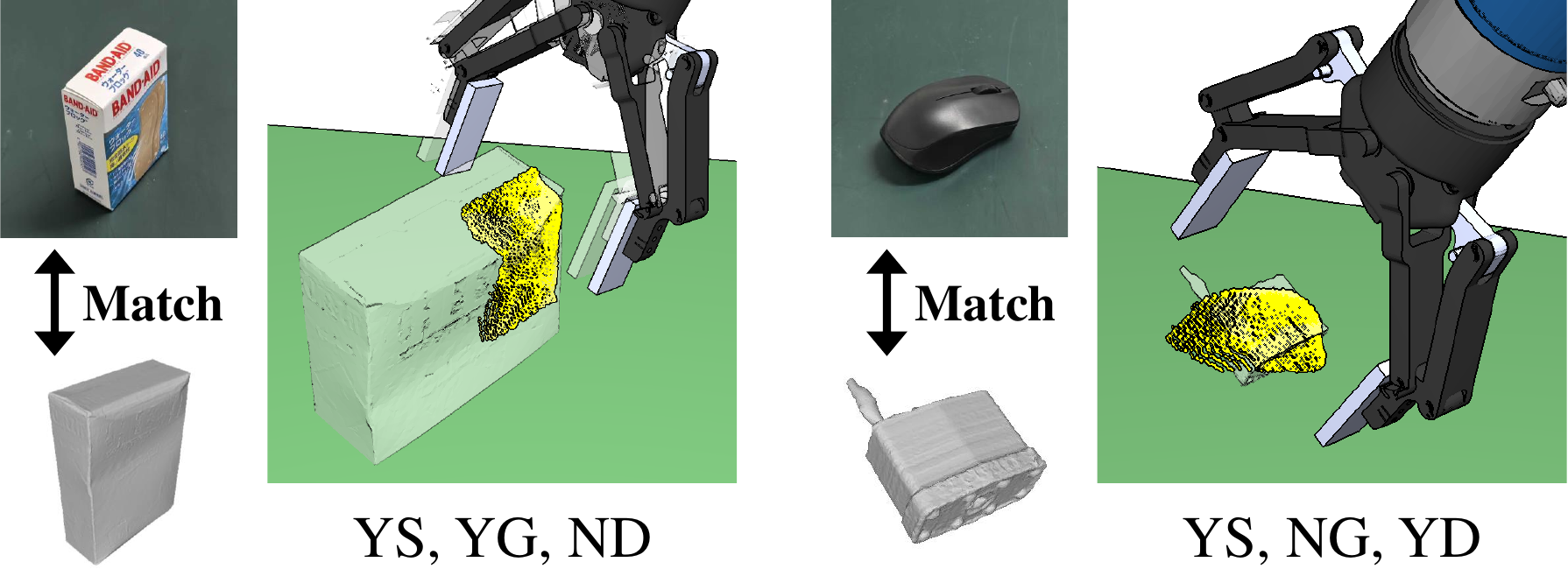}
    \caption{Matching cases where two objects are similar in two out of three perspectives. YS/NS, YG/NG, YD/ND indicate whether (Yes or No) they are Semantically, Geometrically, or Dimensionally similar, respectively.}
    \label{img:17}
\end{figure}

In Section \ref{selection}, we define the candidate model selection rule to prioritize models that exhibit similarity across more perspectives. This strategy outperforms other alternatives, such as using a composite scoring function, by fully exploiting the complementary nature of multi-level similarities. As shown in Fig. \ref{img:17}, a small box (target) is matched with a large box (database model), in which case the two objects are semantically and geometrically similar, but differ in dimension. Despite the size discrepancy, their point clouds align effectively using the PDM-based point cloud registration method, and the planned grasp is subsequently refined through our fine-tuning process, resulting in a robust grasp execution. In another example, a computer mouse (target) is matched with a padlock (database model), which shares semantic and dimensional similarity but differs geometrically, also yielding a successful grasp. These examples demonstrate how the proposed multi-level matching framework facilitates flexible and reliable candidate selection, while the specialized registration and fine-tuning processes ensure high-quality final grasps.

\begin{figure}[t]
    \centering
    \includegraphics[width=\linewidth]{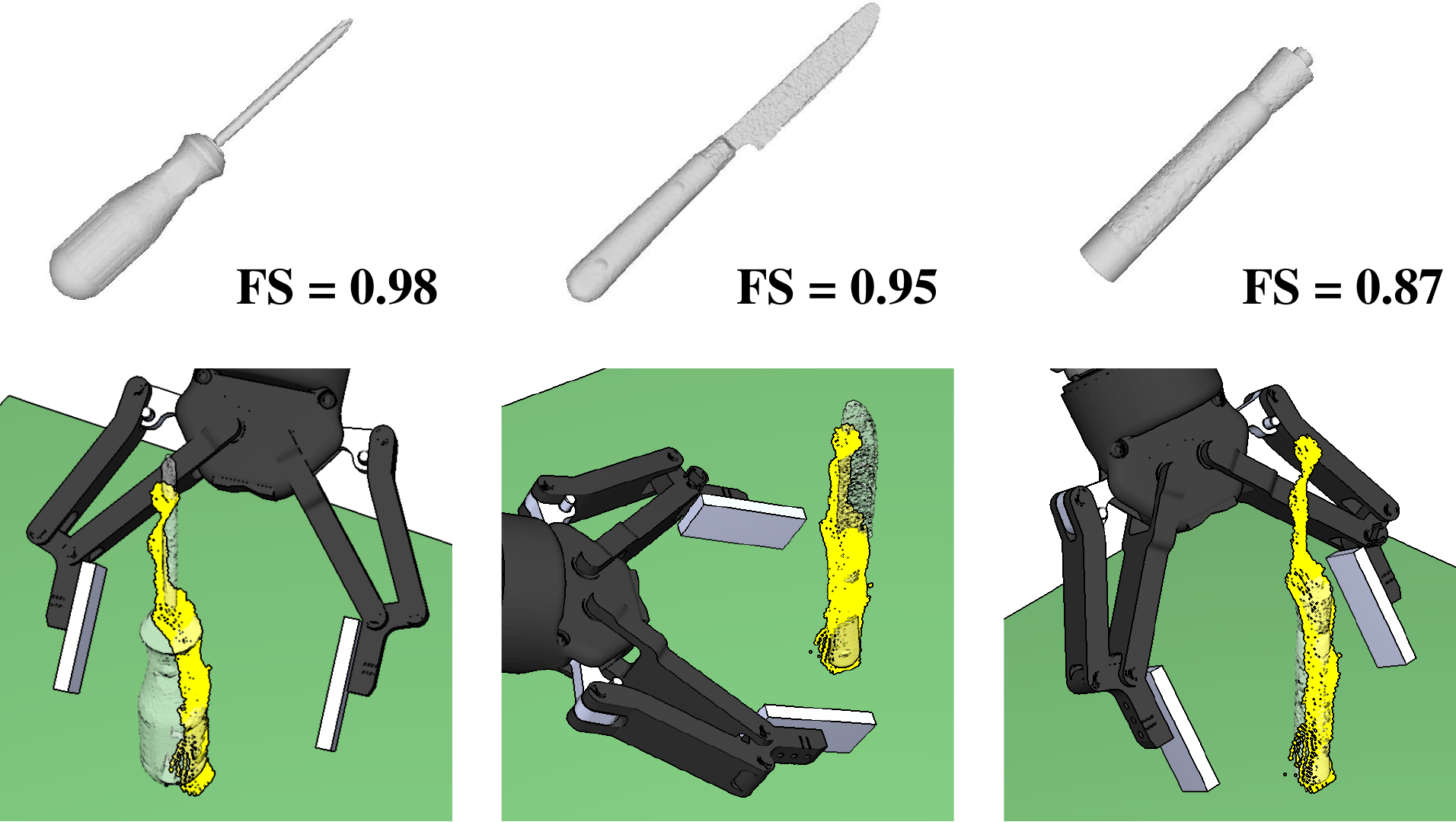}
    \caption{Using different models from the top candidates with high fitness scores to plan grasps for an unknown object (a toothbrush).}
    \label{img:18}
\end{figure}

In Section \ref{ranking}, we rank the selected candidates based on the fitness score (denoted as $\rm FS$) of point cloud registration, which measures the overlap between two aligned point clouds. During grasp planning, our strategy is to use the candidates in order of their $\rm FS$ values, from highest to lowest. However, we find that in most cases, all top candidates with high $\rm FS$ values (typically greater than 0.8) can be effectively used to generate robust grasps. Fig. \ref{img:18} illustrates an example of using different models to generate grasps for the same object. It is evident that all selected models with $\rm FS>0.8$ align well with the target object and yield high-quality grasps. The mere difference is that models with lower $\rm FS$ may only partially resemble the target object (see Fig. \ref{img:18} right), with the unmatched region ignored during grasp planning. While this could result in some loss of information, our experiments show that it has minimal impact on the final planning results, alleviating concerns about the difficulty of determining the order of use when candidate models have similar $\rm FS$ values.

\subsection{Key Feature Extraction of FPFH Descriptors}
\begin{figure}[t]
    \centering
    \includegraphics[width=0.9\linewidth]{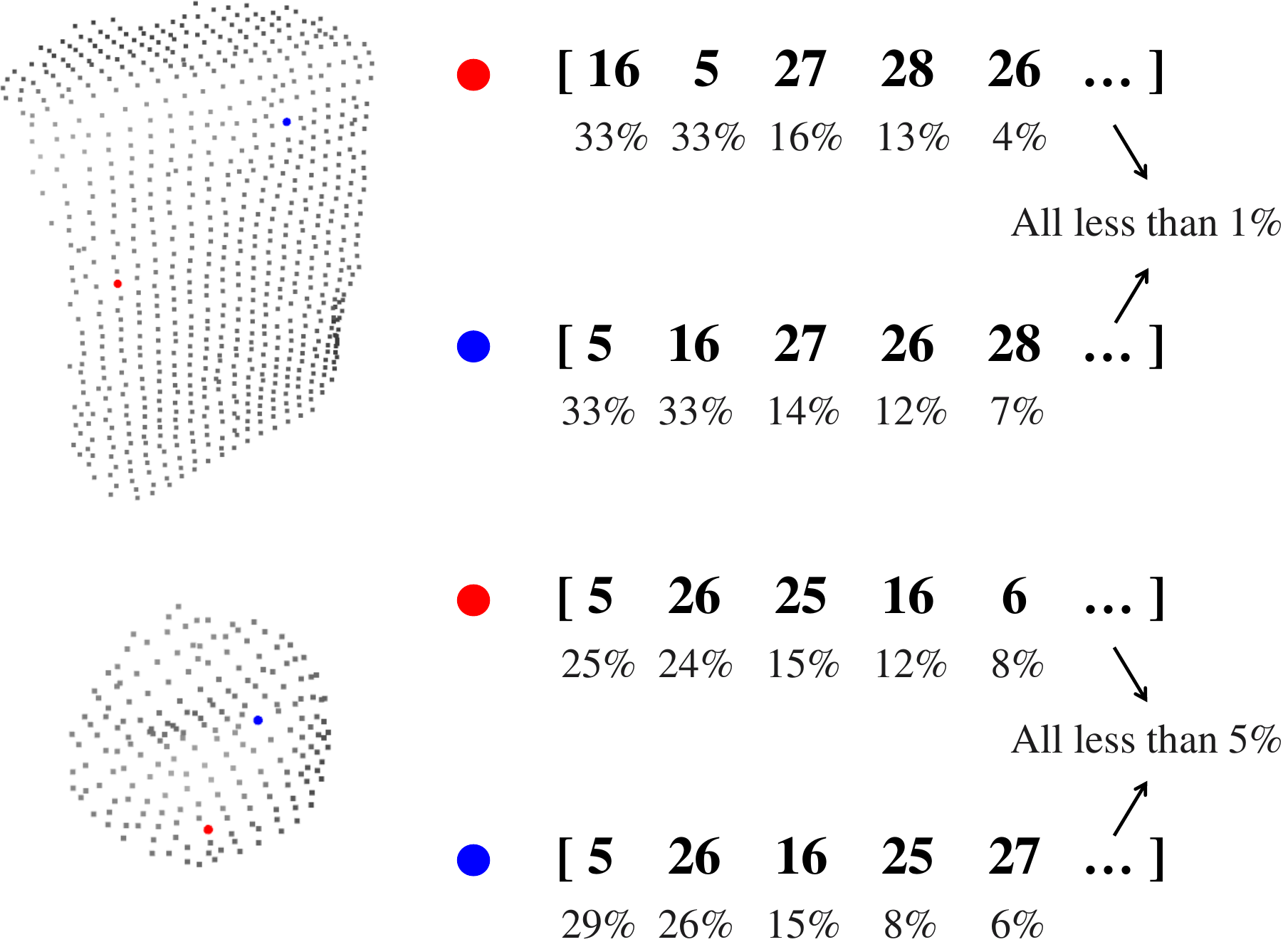}
    \caption{Feature extraction of sampled points from different object regions, with the index number of each FPFH component shown in brackets and their corresponding proportions (ordered from largest to smallest) listed below. It is evident that retaining the first two components ensures optimal robustness.}
    \label{img:19}
\end{figure}

In Section \ref{FPFH}, we propose C-FPFH descriptors by extracting the two most dominant components from the original FPFH descriptors to represent the local geometry of sampled points. This strategy demonstrates strong robustness to sensing noise. As illustrated in Fig. \ref{img:19}, we select two sampled points (shown in red and blue) from the observations of a box and a ball, respectively, and compute their FPFH descriptors. The index numbers of the 33 FPFH components are presented in descending order of their proportions. Although the selected two points are spatially separated, they lie on the same object surface, which means their FPFH descriptors should ideally be identical. However, due to sensing noise, small errors in local geometry can lead to slight variations in the computed FPFH components. Despite this, we observe that the two most dominant components remain consistent between the points, with a clear gap separating them from the remaining components. We hypothesize that this gap arises because points on regular surfaces (e.g., planes or spheres) tend to exhibit similar angular differences in surface normals with their neighbors, resulting in concentrated FPFH component distributions. In contrast, we find that this proportion gap becomes less pronounced—and in some cases, the principal components even vary—for points on complex surfaces. This variability poses a potential limitation when applying FPFH descriptors to objects with intricate geometries. Nevertheless, our multi-level matching framework mitigates this issue by incorporating additional similarity cues beyond local geometric features.

\subsection{Effect of Sensing Noise on Geometric Matching}
To further demonstrate the robustness of C-FPFH-based geometric matching under varying sensing conditions, we conduct a noise effect study by applying two types of depth filters during visual perception: a \textit{spatial filter} which smooths the depth data by considering neighboring pixels, and a \textit{hole-filling filter} which fills in missing depth values by copying nearby valid regions. Both filters are applied using the default parameters provided by the RealSense software.

\begin{figure}[t]
    \centering
    \includegraphics[width=\linewidth]{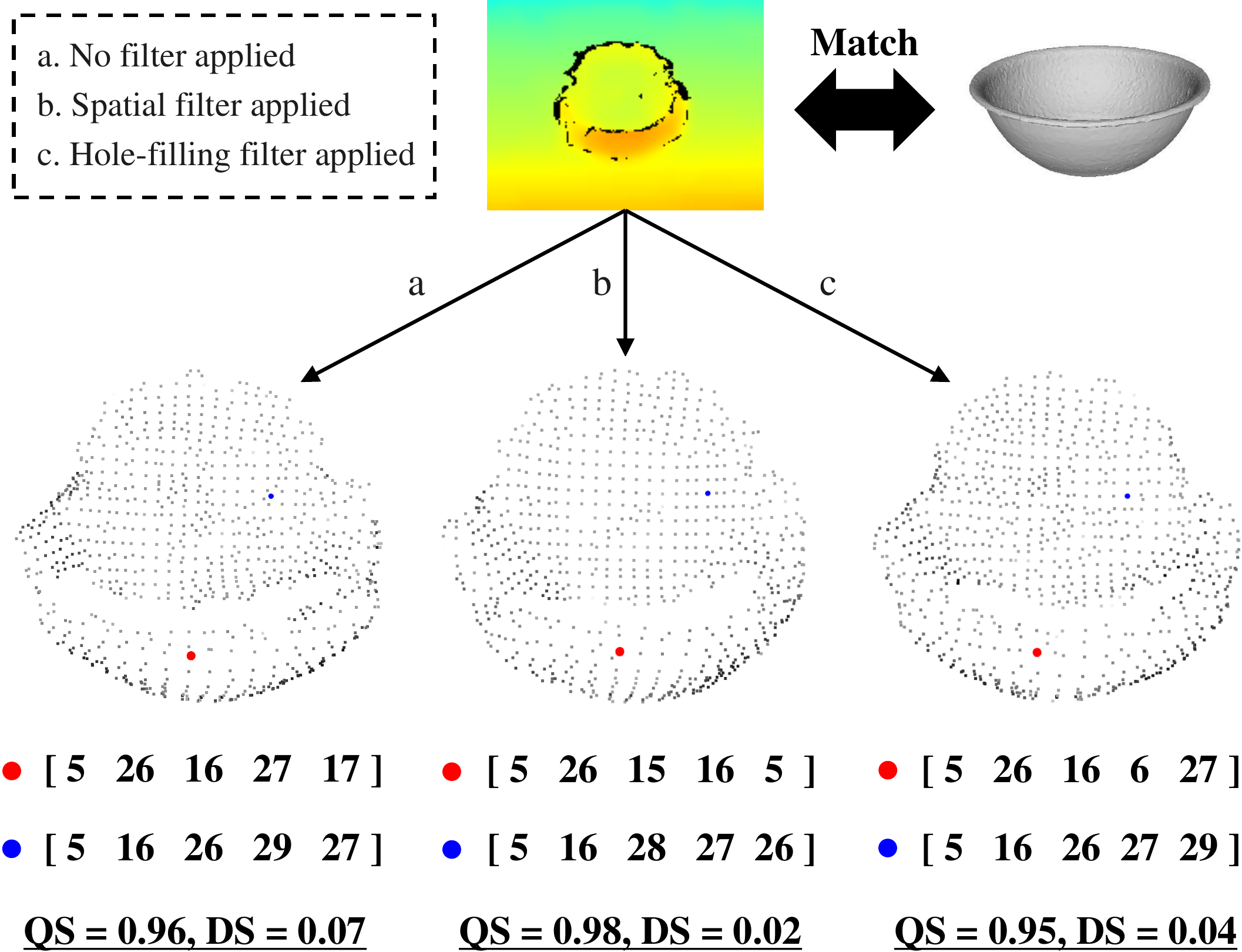}
    \caption{Validations of the robustness of C-FPFH-based geometric matching under varying levels of sensing noise, using different types of depth filters.}
    \label{img:20}
\end{figure}

As shown in Fig. \ref{img:20}, we first obtain a single-view point cloud of a bowl, and then apply the two filters to generate two additional processed point clouds, simulating different levels of sensing noise. To evaluate the impact of noise on local geometry, we select two points (shown in red and blue) at the same spatial locations across the three point clouds and compute their FPFH components. The results show that while the lower-ranked components vary, the two most dominant components remain consistent across all three cases. To verify the effect on overall geometry, we further compute the C-FPFH descriptors for the entire point clouds and match them with a similar database model (also a bowl). The two evaluation metrics, QS and DS, exhibit slight variations but consistently satisfy the thresholds of $\rm QS>0.9$ and $\rm DS<0.1$, in which case the matching results are always \textit{highly similar}. These findings confirm the noise-resistant capability of our geometric matching approach. However, it should also be noted that our task does not address cases involving high levels of noise that significantly alter local geometry.

\section{}
In this appendix, we present additional experiments to assess the generalizability of our approach in two challenging scenarios: grasping objects on non-flat surfaces and handling adversarial objects with minimal similarity to any database models. Furthermore, we validate the scalability of our method by applying it to a larger dataset containing over 1,000 object models across 50 distinct classes, leveraging an optimized candidate selection strategy.

\subsection{Grasping Objects on Non-Flat Surfaces}
\begin{figure}[t]
    \centering
    \includegraphics[width=\linewidth]{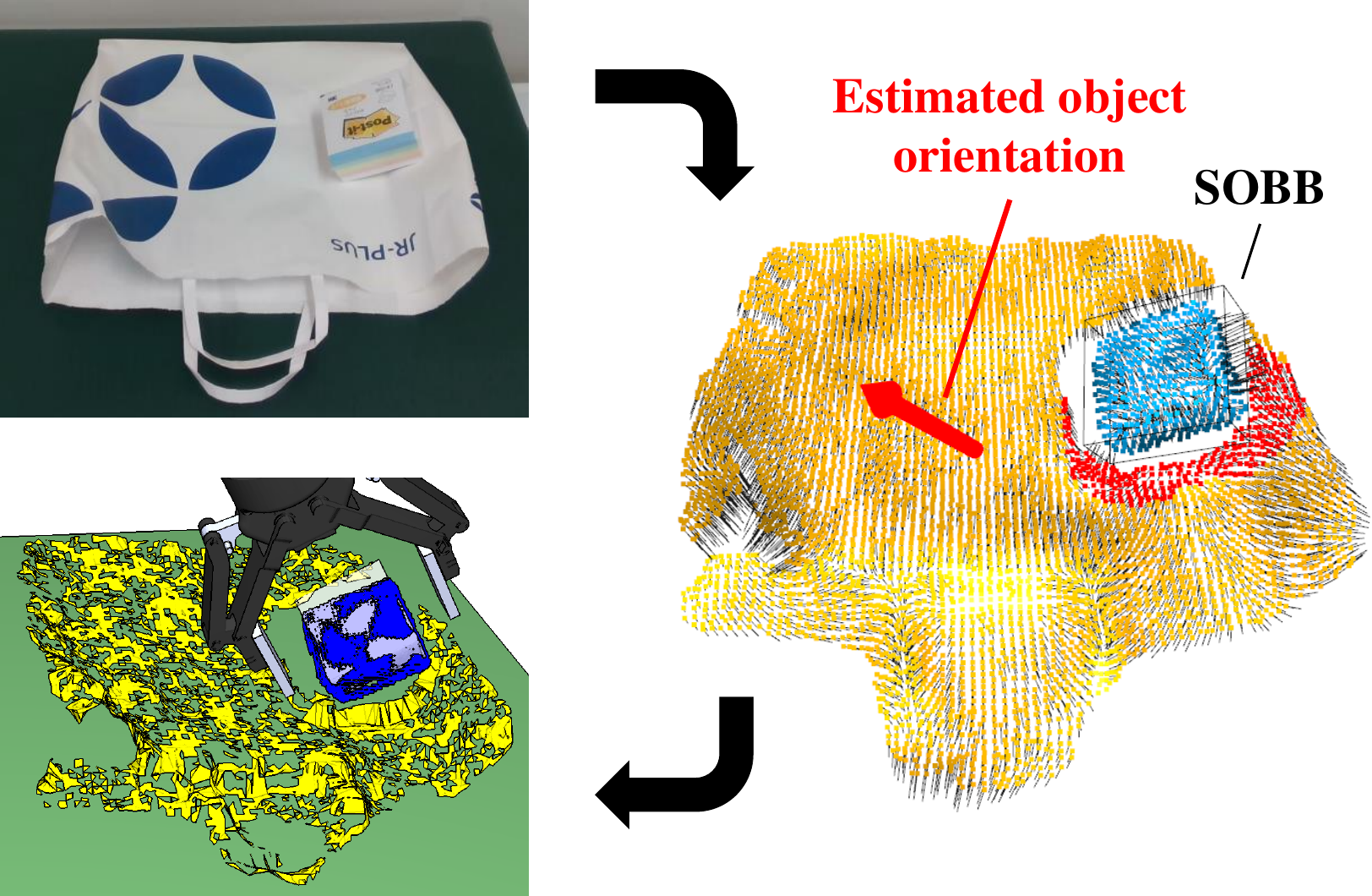}
    \caption{SOBB generation and grasp planning for objects on non-flat surfaces.}
    \label{img:21}
\end{figure}

In Section \ref{SOBB}, we introduce SOBB for accurate evaluation of dimensional similarity. During the SOBB generation process, we need to obtain the normal vector of the plane on which the object is placed. In practice, this normal vector can also be determined for objects resting on non-flat surfaces. As shown in Fig. \ref{img:21}, a small box is placed at an angle on a tote bag with a folded surface. We first extract the point clouds of both the box and the bag (denoted as $p_a$ and $p_b$, respectively) using segmentation methods, followed by surface normal estimation for each. Next, we search for points in $p_b$ whose distance $d$ from the closest point in $p_a$ is below a threshold. In our task, we select points with $d < 2$ cm, which are highlighted in red in Fig. \ref{img:21}. For these selected points, we calculate the average of their normal vectors after removing outliers. This averaged normal vector is then used to estimate the orientation of the object. Based on this orientation, we finally generate the SOBB for the tilted object and perform subsequent similarity matching and grasp planning to successfully complete the task.

However, this estimation method relies on a key assumption: the supporting surface must fully cover the base of the object and its surrounding area must be entirely visible. In other words, SOBB-based matching is not applicable to scenarios such as bin picking, where the support relationships between objects are typically complex and occluded. In such cases, our core approach can still be applied by excluding dimensional matching or by incorporating pose estimation techniques.

\subsection{Grasping Adversarial Objects}
\begin{figure}[t]
    \centering
    \includegraphics[width=0.95\linewidth]{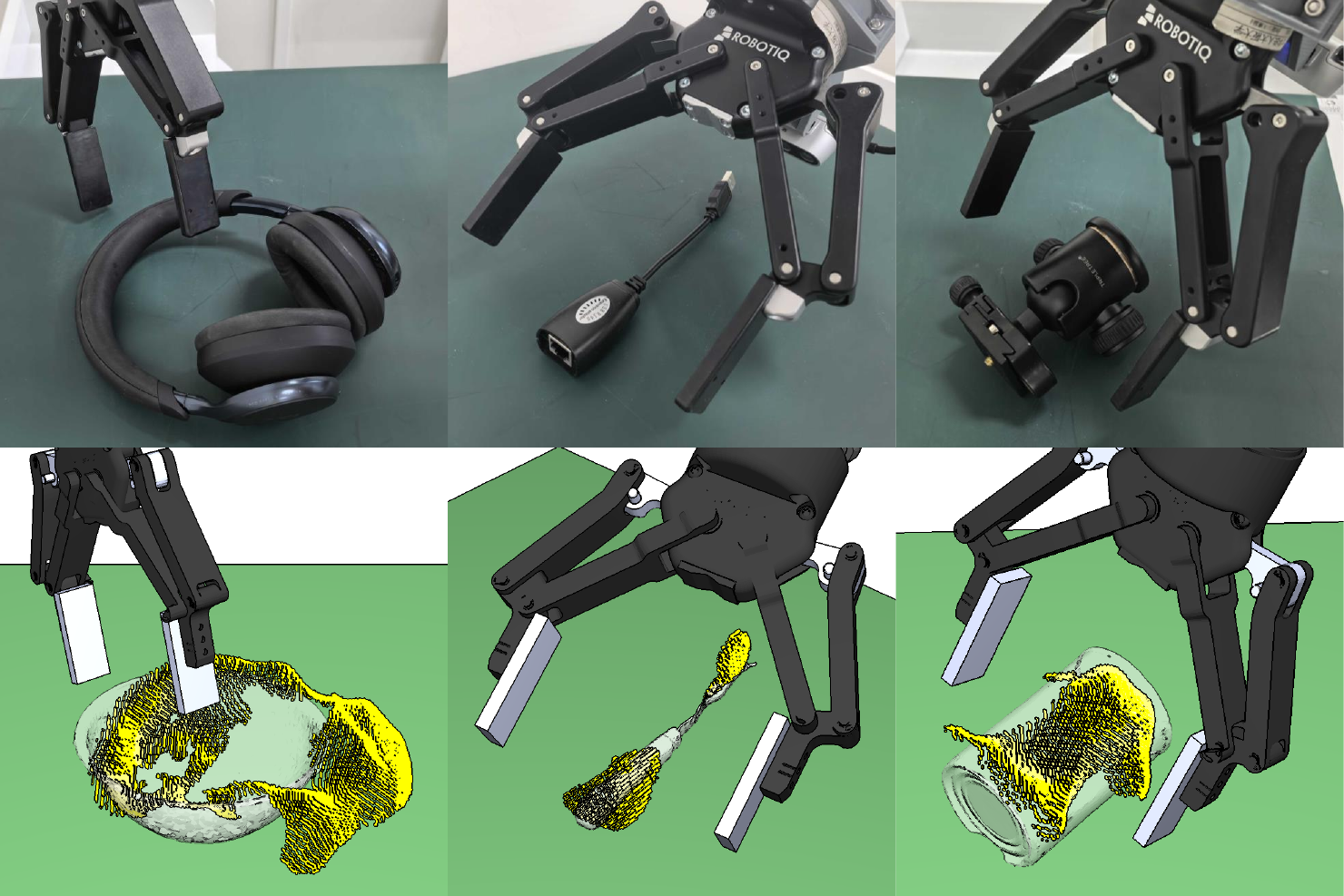}
    \caption{Grasping adversarial objects using our similarity-based approach.}
    \label{img:22}
\end{figure}

In Section \ref{single}, we conduct grasping experiments on a variety of novel objects for which reasonably similar references can be found from the database. To further demonstrate the applicability of our approach to objects that are intuitively dissimilar to any database model, we extend our evaluation to novel-category objects with adversarial geometries. As shown in Fig. \ref{img:22}, we select a headphone, an adapter cable, and a camera gimbal as test cases. The effectiveness of our similarity-based approach is clearly reflected in the matching and planning results. A bowl model, whose edge resembles the frame of the headphone; a fork model, whose handle resembles the adapter port; and a can model, which shares an overall cylindrical shape with the gimbal, are all accurately matched and result in high-quality grasps. This form of \textit{partial similarity} observed in the matching process significantly enhances the generalizability of our method to objects with diverse shapes.

In a comparative experiment with a baseline method (see Table \ref{tab:6}), our approach substantially outperforms GraspNet, achieving a task success rate of 90\% compared to 40\% (a task is considered successful when a valid grasp is planned and executed successfully). The most notable differences appear in the cases of the adapter and the gimbal, where GraspNet generates only sparse and low-quality grasps due to their irregular geometries. This further verifies the superiority of our similarity-based approach over conventional learning-based approaches. However, we also observe that when a suboptimal matching result with a low degree of \textit{partial similarity} occurs, few feasible grasps can be identified during planning, increasing the likelihood of using a \textit{potential grasp} (without quality evaluation). This raises the risk of grasp failure and highlights a potential limitation of the current approach. In future work, we aim to develop a continuously updating framework that incorporates knowledge of novel-category objects, allowing for the renewal of the existing database when failures are detected and providing improved solutions for subsequent handling.

\begin{table}[t]
\small
\renewcommand\arraystretch{1.5}
\setlength\tabcolsep{5pt}
\centering
\caption{Grasping Experiments on Adversarial Objects}
\begin{tabular}{c | c c c | c}
\textbf{Success/Total} & Headphone & Adapter & Gimbal & Average \\
\hline
GraspNet & 10/10 & 0/10 & 2/10 & 12/30 (40\%) \\
Our method & 10/10 & 8/10 & 9/10 & \textbf{27/30 (90\%)} \\
\end{tabular}
\label{tab:6}
\end{table}

\subsection{Validation on Large-Scale Datasets}
In the experiment section, we use the YCB dataset consisting of fewer than 100 object models for similarity matching. To further verify the scalability of our approach, we leverage a larger dataset from Dex-Net, which contains over 1,000 object models across 50 distinct classes. However, we observe that the model selection strategy needs refinement to effectively suppress computation time as the number of models grows. Instead of performing similarity matching on the entire dataset, we first extract a subset of models and restrict further matching to this reduced set. To support this extraction process, we develop two pre-filtering strategies from the perspectives of semantic and dimensional similarity, respectively.

For semantic pre-filtering, we utilize the Word2Vec word embedding approach to compute the cosine similarity between the category name of the target object and those of the database models. The resulting similarity scores range from 0 to 1, as shown in Table \ref{tab:7}. As the score distributions vary across different categories, we use a ratio $\delta$ (set to 50\% in our task) rather than a fixed cutoff to filter out object categories with low similarity. This ratio is adjustable for balancing the trade-off between broader exploration and lower computational load. The key advantage of Word2Vec in this context is its ability to efficiently compute similarity scores for a large number of word pairs simultaneously, thereby compensating for the weaknesses of LLMs, such as limited prompt length and reduced performance when handling numerous object~categories.

\begin{table}[t]
\small
\renewcommand\arraystretch{1.5}
\setlength\tabcolsep{5pt}
\centering
\caption{Word2Vec Cosine Similarity Between Object Categories}
\begin{tabular}{c | c c c c c c c}
\textbf{Score} & Bottle & Bowl & Apple & Banana & Scissor & Hammer \\
\hline
Box & \textbf{0.46} & \textbf{0.36} & 0.24 & 0.23 & 0.26 & 0.30 \\
Pear & 0.36 & 0.31 & \textbf{0.69} & \textbf{0.59} & 0.17 & 0.25 \\
Knife & 0.38 & 0.33 & 0.25 & 0.29 & \textbf{0.55} & \textbf{0.57} \\
\end{tabular}
\label{tab:7}
\end{table}
\begin{figure}[t]
    \centering
    \includegraphics[width=0.95\linewidth]{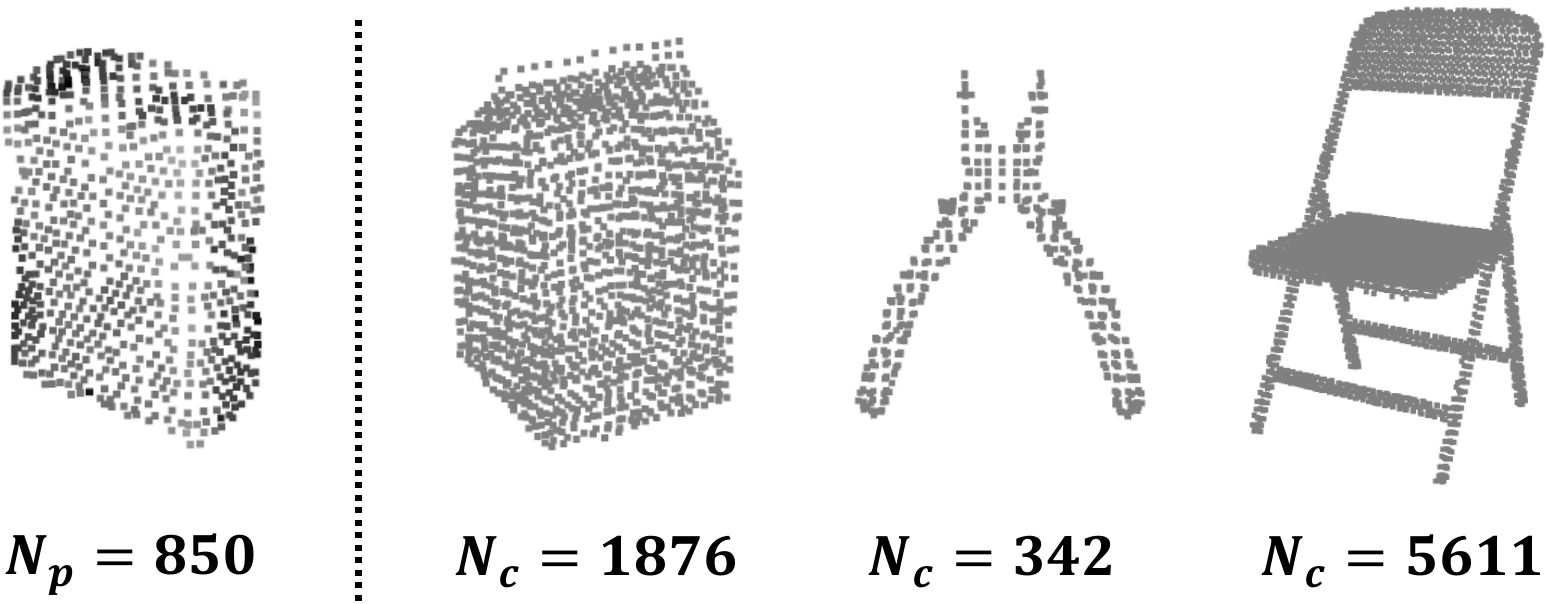}
    \caption{Object point clouds after being downsampled with the same voxel size. $N_p$ and $N_c$ represent the number of points in the observed partial point cloud and the database complete point cloud, respectively.}
    \label{img:23}
\end{figure}

For dimensional pre-filtering, we downsample the point clouds of both the observed object and the database models using the same voxel size (5 mm in our task), recording the number of points in each, denoted as $N_p$ for the observed partial point cloud and $N_c$ for the database complete point cloud. Fig. \ref{img:23} illustrates an example where the target object is a box, alongside a set of database models (milkbox, pliers, chair) varying in the number of points due to size discrepancies. Assuming the target object approximates a cube and a diagonally downward-facing camera can typically observe 2 to 3 of its surfaces, the number of points in a highly similar model should generally be 2 to 3 times that of the observed object. Based on this assumption, we define a range such that only models within $\alpha N_p<N_c<\beta N_p$ ($\alpha=0.5, \beta=5$ in our task) are selected for further matching. In the case shown in Fig. \ref{img:23}, the pliers and the chair models are excluded during the pre-filtering step. The key advantage of voxel downsampling in this context is that simple comparisons of the number of points significantly reduce computational complexity compared to bounding box estimations.

Based on these pre-filtering methods, we conduct grasping experiments on the same set of objects as in Section \ref{single}. The results are comparable, with a 95\% grasp success rate and a 100\% plan success rate. We observe that the increase in the number of models enhances the likelihood of identifying a highly similar reference, showcasing the advantage of using a large dataset. Meanwhile, the planning time only increases by about 1 second compared to using the smaller dataset for computing Word2Vec similarity, demonstrating the consistent high efficiency of our approach via the subset pre-sorting~strategy.

\section*{Acknowledgements}
This research is subsidized by New Energy and Industrial Technology Development Organization (NEDO) under a project JPNP20016. This paper is one of the achievements of joint research with and is jointly owned copyrighted material of ROBOT Industrial Basic Technology Collaborative Innovation Partnership (ROBOCIP).

\bibliographystyle{IEEEtran}
\bibliography{reference}

\vspace{-2.0em}
\begin{IEEEbiography}[{\includegraphics[width=1in,height=1.25in,clip,keepaspectratio]{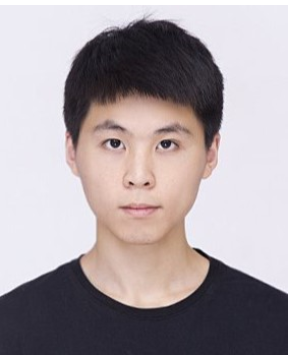}}]{Hao Chen} (Member, IEEE) received the B.Eng. degree from Zhejiang University, China, in 2019, and the M.S. degree from the Graduate School of Engineering Science, Osaka University, Japan, in 2022. He is currently pursuing a Ph.D. in the Graduate School of Engineering Science at Osaka University. His research interests include robotic grasping and computer vision.
\end{IEEEbiography}
\vspace{-2.0em}
\begin{IEEEbiography}
[{\includegraphics[width=1in,height=1.25in,clip,keepaspectratio]{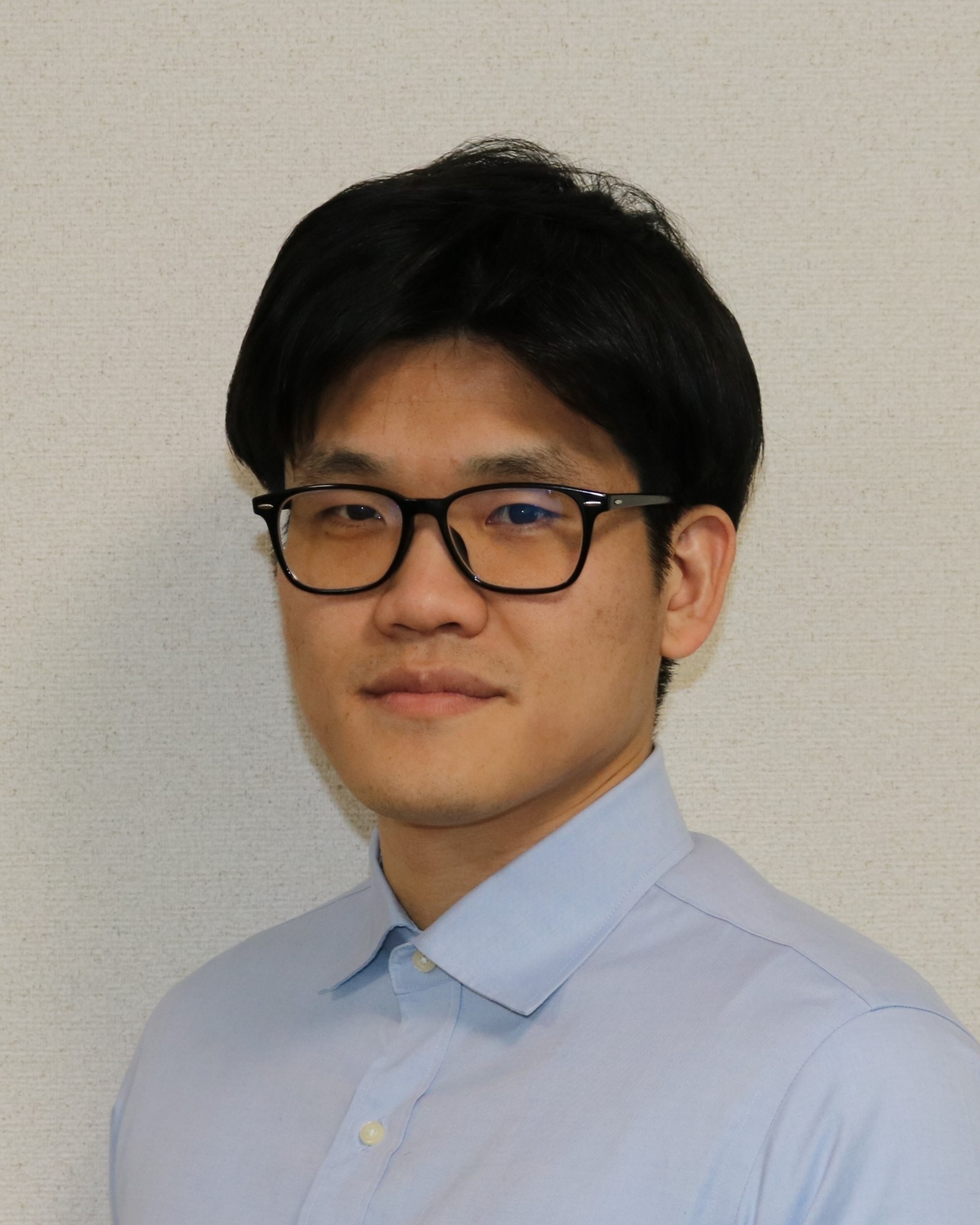}}]{Takuya Kiyokawa} (Member, IEEE) received the Ph.D. degree from the Nara Institute of Science and Technology, Japan, in 2021. From 2021 to 2022, he was with Osaka University and the Nara Institute of Science and Technology, both in Japan, as a Specially Appointed Assistant Professor. Since 2023, he has been an Assistant Professor at Osaka University. From 2023 to 2024, he was a Visiting Researcher at the Institute of Robotics and Mechatronics, German Aerospace Center (DLR), Wessling, Germany. His current research interests include reconfigurable manipulation robots.
\end{IEEEbiography}
\vspace{-2.0em}
\begin{IEEEbiography}
[{\includegraphics[width=1in,height=1.25in,clip,keepaspectratio]{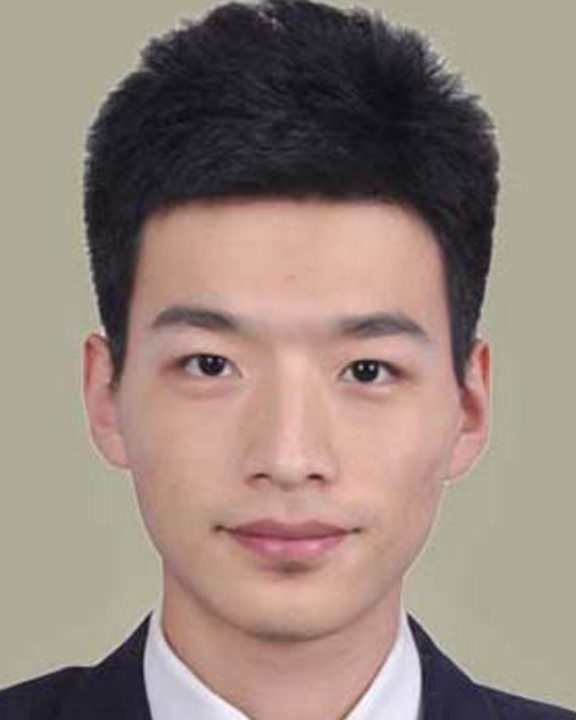}}]{Zhengtao Hu} (Member, IEEE) received the B.Eng. and M.S. degrees in energy power systems and automation from Xi’an Jiaotong University, Xi’an, China, in 2015 and 2018, respectively, and the Ph.D. degree in systems science from the Graduate School of Engineering Science, Osaka University, Suita, Japan, in 2022. From 2022 to 2023, he was a Research Assistant Professor with Osaka University, Suita, Japan. Since 2023, he has been a Lecture with the School of Mechatronic Engineering and Automation, Shanghai University, Shanghai, China. His research interests focus on robotic manipulation, mechanism design, and robotic vision.
\end{IEEEbiography}
\vspace{-2.0em}
\begin{IEEEbiography}
[{\includegraphics[width=1in,height=1.25in,clip,keepaspectratio]{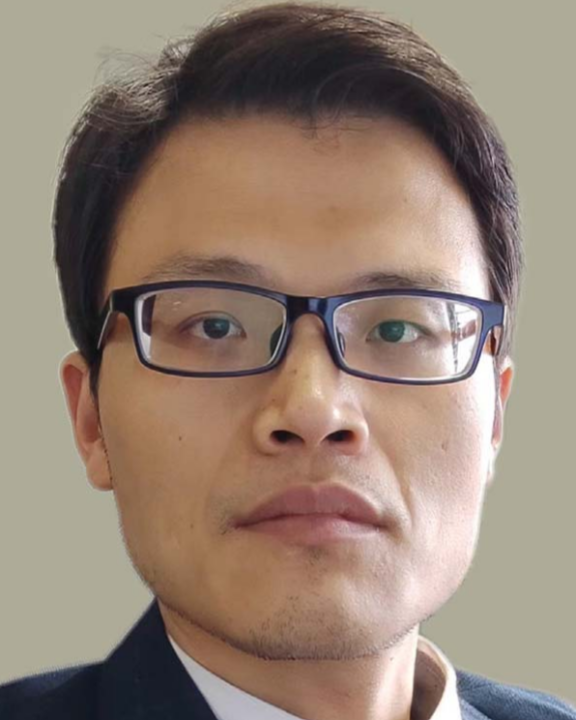}}]{Weiwei Wan} (Senior Member, IEEE) received the Ph.D. degree from the Department of Mechano Informatics, The University of Tokyo, Tokyo, Japan, in 2013. From 2013 to 2015, he did the post doctoral research at Carnegie Mellon University, PA, USA, under the support of Japanese Society for the Promotion of Science (JSPS). From 2015 to 2017, he was a Research Scientist at the National Institute of Advanced Industrial Science and Technology (AIST), Tsukuba, Japan. He is currently an Associate Professor working with the Graduate School of Engineering Science, Osaka University, Japan. His research interests include robotic manipulation and smart manufacturing.
\end{IEEEbiography}
\vspace{-2.0em}
\begin{IEEEbiography}
[{\includegraphics[width=1in,height=1.25in,clip,keepaspectratio]{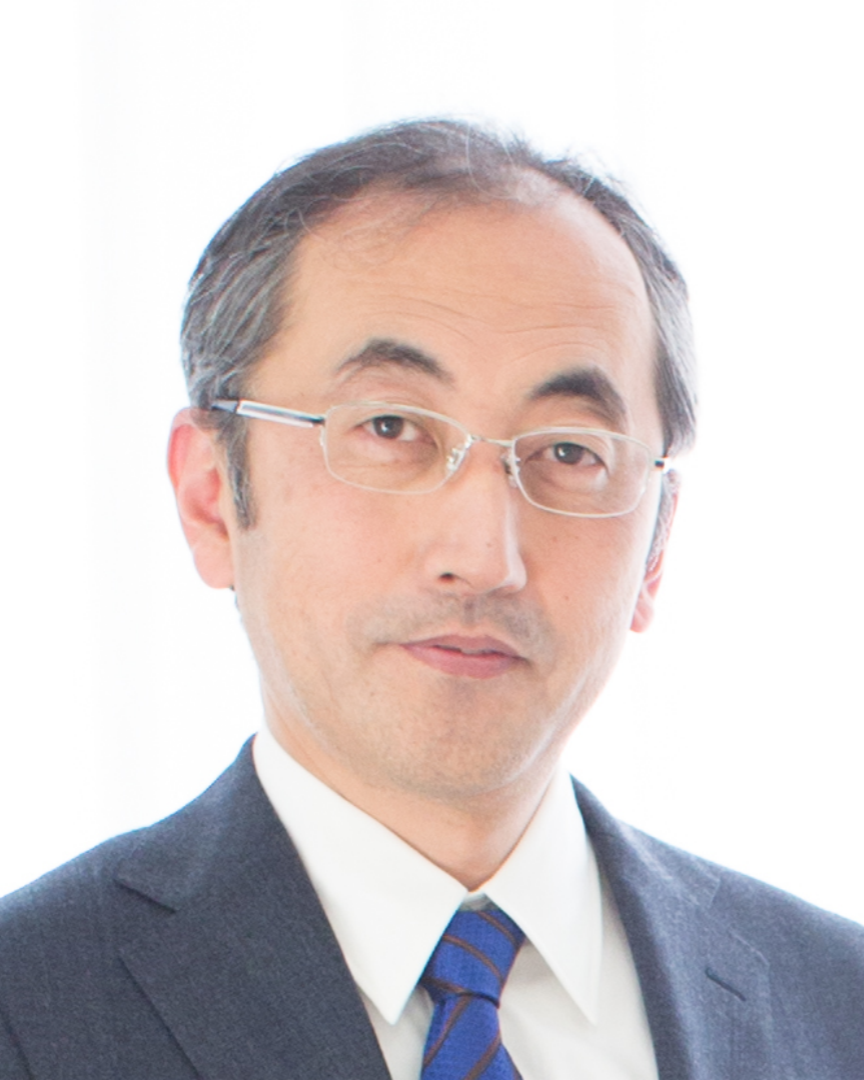}}]{Kensuke Harada} (Fellow, IEEE) received the Ph.D. degree from Kyoto University, Kyoto, Japan, in 1997. From 1997 to 2002, he was a Research Associate at Hiroshima University, Hiroshima, Japan. From 2005 to 2006, he was a Visiting Scholar at the Computer Science Department, Stanford University, California, USA. He was a Research scientist at National Institute of Advanced Industrial Science and Technology (AIST), Tsukuba, Japan. He is currently a Professor working with the Graduate School of Engineering Science, Osaka University. His research interests include the mechanics and planning of robotic systems and its industrial applications.
\end{IEEEbiography}

\end{document}